\documentclass[lettersize,journal]{IEEEtran}
\pdfoutput=1
\usepackage{amsmath,amsfonts}
\usepackage{algorithmic}
\usepackage{algorithm}
\usepackage{array}
\usepackage[caption=false,font=tiny,labelfont=sf,textfont=sf]{subfig}	
\usepackage{textcomp}
\usepackage{stfloats}
\usepackage{url}
\usepackage{verbatim}
\usepackage{graphicx}
\usepackage{cite}
\usepackage{makecell}
\usepackage{multirow}
\usepackage{amsfonts}	
\usepackage{amssymb}	
\newcolumntype{Y}{>{\centering\arraybackslash}X}

\usepackage{color}
\usepackage{bm}

\begin{document}
	
	\title{View-aware Salient Object Detection \\ for 360° Omnidirectional Image}
	
	\author{Junjie~Wu, Changqun~Xia, Tianshu~Yu, Jia~Li, ~\IEEEmembership{Senior Member,~IEEE}
		\thanks{J. Wu, T. Yu and J. Li are with the State Key Laboratory of Virtual Reality Technology and Systems, School of Computer Science and Engineering, Beihang University, Beijing, 100191, China. J. Li is also with Peng Cheng Laboratory, Shenzhen, 518000, China.}
		\thanks{C. Xia is with Peng Cheng Laboratory, Shenzhen, 518000, China.}
		\thanks{Correspondence should be addressed to C. Xia (e-mail: xiachq@pcl.ac.cn) and Jia Li (e-mail: jiali@buaa.edu.cn). Website: http://cvteam.net}
		}
	
	\markboth{Submission to IEEE TRANSACTIONS ON MULTIMEDIA}%
	{Shell \MakeLowercase{\textit{et al.}}: A Sample Article Using IEEEtran.cls for IEEE Journals}
	
	
	\maketitle
	
	\begin{abstract}
		Image-based salient object detection (ISOD) in $360^{\circ}$ scenarios is significant for understanding and applying panoramic information. 
		However, research on $360^{\circ}$ ISOD has not been widely explored due to the lack of large, complex, high-resolution, and well-labeled datasets. 
		Towards this end, we construct a large scale $360^{\circ}$ ISOD dataset with object-level pixel-wise annotation on equirectangular projection (ERP), which contains rich panoramic scenes with not less than 2K resolution and is the largest dataset for $360^{\circ}$ ISOD by far to our best knowledge. 
		By observing the data, we find current methods face three significant challenges in panoramic scenarios: diverse distortion degrees, discontinuous edge effects and changeable object scales. 
		Inspired by humans' observing process, we propose a view-aware salient object detection method based on a Sample Adaptive View Transformer (SAVT) module with two sub-modules to mitigate these issues. Specifically, the sub-module View Transformer (VT) contains three transform branches based on different kinds of transformations to learn various features under different views and heighten the model's feature toleration of distortion, edge effects and object scales. 
		Moreover, the sub-module Sample Adaptive Fusion (SAF) is to adjust the weights of different transform branches based on various sample features and make transformed enhanced features fuse more appropriately. 
		The benchmark results of 20 state-of-the-art ISOD methods reveal the constructed dataset is very challenging. Moreover, exhaustive experiments verify the proposed approach is practical and outperforms the state-of-the-art methods.
	\end{abstract}
	
	
	\section{Introduction}
	\IEEEPARstart{O}{mnidirectional} images can sample the entire viewing sphere surrounding its optical center, a $360^{\circ}\times180^{\circ}$ FoV~\cite{su2017learning,zhang2020fixation} and the resolution of an omnidirectional image (ODI) is always several times that of the traditional image, making storing, transmitting and understanding more difficult~\cite{ng2005data}. Therefore, salient object detection, automatically processing regions of interest and selectively ignoring parts of uninterest, is significant for compressing, transmitting and analyzing $360^{\circ}$ panoramic images~\cite{xu2020state,de2017look,li2017closed}.
	
	\begin{figure}[!t]
		\centering
		\includegraphics[width=0.98\columnwidth]{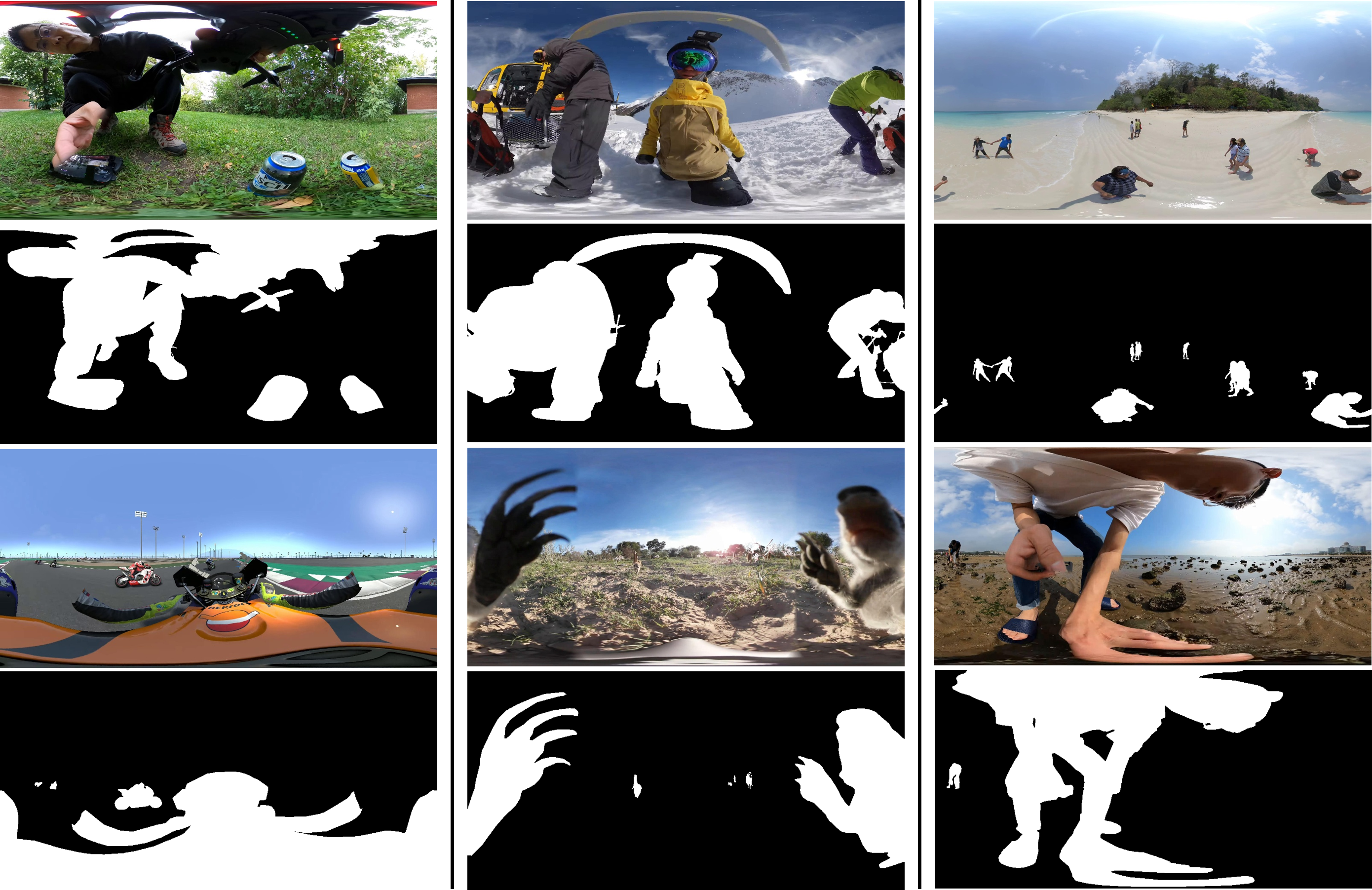} 
		\caption{Representative examples of $360^{\circ}$ omnidirectional images in ERP form. These three columns show the cases of diverse distortion degrees, discontinuous edge effects and changeable object scales, respectively. }
		\label{fig:introduction}
	\end{figure}
	Currently, panoramic datasets are evolving to meet increasing demands in benchmarking and developing $360^{\circ}$-based ISOD models. Li et al.~\cite{li2019distortion} construct the first $360^{\circ}$ ISOD dataset 360-SOD with pixel-wise object-level annotation containing 500 equirectangular projection (ERP) images in $512\times1024$ resolution from existing human fixation datasets. Ma et al.~\cite{ma2020stage} collect 1105 ERP images in $546\times1024$ resolution with object-level annotation. Zhang et al.~\cite{zhang2020fixation} provide a dataset with object-level and instance-level annotation containing 107 ERP images in $512\times1024$ resolution.
	However, currently available datasets are relatively small in scale and resolution and less complex in scenarios, which is not enough for further studies and is a primary cause for limited related research. Besides, insufficient training data easily leads to model overfitting. Therefore, it is urgent to break the data bottleneck.

	To this end, we construct a new large scale $360^{\circ}$ omnidirectional image-based salient object detection (SOD) dataset referred to as ODI-SOD with object-level pixel-wise annotation on ERP to assist studies about $360^{\circ}$ ISOD task. The proposed dataset contains 6,263 ERP images with not less than 2K resolution selected from 8,896 panoramic images and 998 videos. The chosen images have the number of salient regions ranging from one to more than ten, the area ratios of salient regions from less than $0.02\%$ to more than $65\%$ and the resolutions from 2K to 8K. More than half of the scenarios are complex and contain diverse objects.

	Moreover, through the observation of the dataset, we find that the poor performances of existing state-of-the-art methods~\cite{li2019distortion, xu2021locate,ZoomNet-CVPR2022} can attribute to three prominent challenges, i.e., diverse distortion degrees, discontinuous edge effects and changeable object scales. Distortion varying with the projection position (e.g., the first col in Fig.\ref{fig:introduction}) leads to uniform filter sampling and feature learning difficulties. Edge discontinuity (e.g., the animal is split into two parts on the borders in the second col of Fig.\ref{fig:introduction}) makes it difficult to segment complete salient objects on projection borders. Changeable scale objects, especially small/large ones in wide FoV panoramas (e.g., the last cols in Fig.\ref{fig:introduction}), make detection and segmentation more difficult. 	
	
	In fact, to mitigate above challenges in $360^{\circ}$ ISOD, researchers have made some related attempts. For example, Li et al.\cite{li2019distortion} propose a distortion-adaptive module that 
	cuts each ERP image into four image blocks to learn different kernels and design a multi-scale module to integrate context features. Ma et al.\cite{ma2020stage} put forward a multi-stage ISOD method to handle ERP distortions by using less distorted perspective images and object-level semantical saliency ranking. However, these methods mainly focus on alleviating distortion, ignoring that an advantageous characteristic of a panorama is the continuous complete panoramic field of view. Cutting an ERP image into blocks makes a panorama lose its full panoramic view and may only segment out partial object regions due to complete objects being broken. Perspective images also contain a limited field of view and may be affected by the previous stage's mistakes. Moreover, multi-stage learning and perspective images demanding heavy memory are not friendly to model training/testing. 	
	
	The FoV of a panorama is $360^{\circ}\times180^{\circ}$ while the binocular visual field of human beings is about $120^{\circ}$~\cite{wiki_FoV}. To better understand the panorama, humans usually change the viewpoint (e.g., look up/down or left/right) or adjust view distance (e.g, zoom in/out) to obtain more scene information from different perspectives.  
	The whole observing process is smooth and keeps the complete panoramic view.  
	Inspired by the observing behaviors of human beings, in this paper  
	we put forward a solid Sample Adaptive View Transformer (SAVT) module based on various geometric transformations.
	SAVT contains two sub-modules, View Transformer (VT) and Sample Adaptive Fusion (SAF). Simulating humans' observing process, VT makes different feature transformations based on different ERP center viewpoints or view distances to learn various features under different views. 
	Following VT, SAF generates adaptive weights for different transform branches based on sample characteristics and makes the features fuse better.  
	Combining VT and SAF, the effects of SAVT are threefold: 1) mitigating the effects of discontinuous edges by changing center viewpoints, 2) better locating and segmenting objects in changeable scales by converting view distances to obtain scalable scene information, and 3) heightening the feature toleration of distortion by increasing the distortion diversity. 
	It is different from the methods of adapting distortion or reducing distortion by adjusting filter sampling methods~{\cite{fernandez2020corners,tateno2018distortion}}. 
	Benchmark results on 20 state-of-the-art ISOD methods present the proposed dataset is challenging. Moreover, qualitative and quantitative experiments verify the proposed method is effective and outperforms the state-of-the-art methods.
	
	Our contributions are as follows:
	\begin{itemize}
		\item \textbf{We construct a new large-scale challenging \bm{$360^{\circ}$} ISOD dataset named ODI-SOD. It contains 6263 high-resolution ERP images with object-level pixel-wise annotation and is the largest \bm{$360^{\circ}$} ISOD dataset to the best of our knowledge. }
		\item \textbf{To our best knowledge, we should be the first one to transfer humans' observing process for panoramas to deep feature learning for ERP images.}
		\item \textbf{Inspired by humans' observing behaviors, we propose a novel Sample Adaptive View Transformer (SAVT) module, which keeps the complete panoramic view and mitigates the effects of distortion, edge discontinuity, and changeable scale objects in panoramic scenarios.} 
		\item \textbf{We make a benchmark on the proposed dataset using 2D ISOD methods, \bm{$360^{\circ}$} panoramic ISOD methods and our methods. Our approach outperforms existing state-of-the-art methods.}
	\end{itemize}

	\section{Related Work}
	In this section, we briefly review existing mainstream $360^{\circ}$ panoramic datasets and $360^{\circ}$ panoramic models. 
	\subsection{$360^{\circ}$ Panoramic Datasets}	
	Datasets play an important role in object detection tasks such as salient object detection~\cite{yan2013hierarchical_ecssd}, co-salient object detection~\cite{fan2020taking}, RGB-D salient object detection~\cite{fan2020rethinking, zhou2021rgb} and camouflaged objects detection\cite{fan2021concealed, fan2020camouflaged}.
	For example, in the 2D domain, the remarkable progress of the ISOD task benefits much from the construction of representative datasets~\cite{yan2013hierarchical_ecssd,yang2013saliency_dutomron,li2014secrets_pascals,cheng2014global_MSRA10K,li2015visual_hkuis,xia2017and_xpie,wang2017learning_duts,li2017instance_ilso1k,fan2018salient_soc,fan2020rethinking_sip,li2021instance_ilso2k}. Early datasets are often limited in the number of images or scene complexity~\cite{yan2013hierarchical_ecssd, yang2013saliency_dutomron, li2014secrets_pascals,cheng2014global_MSRA10K,li2015visual_hkuis}. Whereafter, 
	two large-scale and challenging datasets XPIE~\cite{xia2017and_xpie} and DUTS~\cite{wang2017learning_duts} are introduced to overcome preceding shortcomings. Besides, salient object datasets with instance-level annotation are proposed~\cite{li2017instance_ilso1k,fan2018salient_soc,fan2020rethinking_sip,li2021instance_ilso2k} to promote the research.

	Recently, some researchers~\cite{rai2017dataset,cheng2018cube,xu2018gaze,zhang2018saliency, yang2018object,fang2018novel,zhao2020spherical, li2019distortion,zhang2020fixation,ma2020stage, zhang2021shd360,zhang2021asod60k,zhou2021omnidirectional,chao2020multi,fang2022database,fang2022perceptual} turn attention to saliency studies in $360^{\circ}$ panoramic scenarios. While most datasets only provide either eye-fixation groundtruth data for saliency prediction or bounding box groundtruth for object detection, which can promote salient object detection but is not enough for accurate pixel-wise salient object segmentation in panoramic scenarios. Therefore, three small-scale omnidirectional image-based SOD datasets with pixel-wise annotation, i.e., 360-SOD~\cite{li2019distortion}, F-360iSOD~\cite{zhang2020fixation} and 360-SSOD~\cite{ma2020stage}, are successively proposed for $360^{\circ}$ ISOD. 
	360-SOD~\cite{li2019distortion} is the first $360^{\circ}$ SOD dataset and has 500 ERP images with pixel-wise object-level annotation based on human fixation groundtruth. F-360iSOD~\cite{zhang2020fixation} is the first $360^{\circ}$ SOD dataset providing pixel-wise object level and instance level binary masks and contains 107 ERP images with 1,165 salient objects. The latest dataset 360-SSOD~\cite{ma2020stage} has 1,105 semantically
	balanced ERP images with only object-level masks. 
	To our best knowledge, they are the datasets available for the $360^{\circ}$ ISOD task.
	
	However, the available datasets are insufficient in number or the scene complexity to understand the real-world panoramic scenarios. 
	It is expected that a large-scale high-resolution dataset with rich and complex scenarios is built to alleviate data constraints. Rich and complex scenarios are closer to the real world, the large-scale number is helpful for training models, and the high-resolution represents the detail information better. Thus, in this paper, we introduce a large-scale $360^{\circ}$ ISOD dataset with high-resolution and complex scenarios. The general information of representative datasets for 2D ISOD and $360^{\circ}$ panoramic ISOD is shown in Tab.I.

	\begin{table*}[!h]
		\centering{
			\caption{Representative datasets for ISOD.}
			\begin{tabular}{c|c|c|c|c|c|c|p{0.3\textwidth}<{\centering}}
				\hline
				Task & Dataset 	& Year & \#Image & \#GT & Res.[min, max] & GT Level & Description  \tabularnewline
				\hline
				&ECSSD~\cite{yan2013hierarchical_ecssd} &CVPR’13 &1,000 &1,000 &[139, 400] &obj. & \makecell*[c]{includes many semantically meaningful \\but structurally complex images}    \tabularnewline  
				\cline{2-8}
				&DUT-OMRON~\cite{yang2013saliency_dutomron} &CVPR’13 &5,168 &5,168 &[139, 401] &obj. &\makecell*[c]{one or more salient objects and \\relatively complex background}    \tabularnewline  
				\cline{2-8}					
				&PASCAL-S~\cite{li2014secrets_pascals} &CVPR’14 &850 &850 &[139, 500] &obj. &\makecell*[c]{multiple salient objects, \\total 1296 relatively salient object instances}  \tabularnewline 
				\cline{2-8}
				\multirow{11}{*}{\rotatebox{90}{2D ISOD}} 
				&MSRA10K~\cite{cheng2014global_MSRA10K} &TPAMI’14 &10,000 &10,000 &[165, 400] &obj. &\makecell*[c]{most are with only one salient \\object and simple background.}   \tabularnewline 
				\cline{2-8}
				&HKU-IS~\cite{li2015visual_hkuis} &CVPR'15 &4,447 &4,447 &[100, 500] &obj. &\makecell*[c]{most have either low contrast, complex \\background or multiple salient objects}   \tabularnewline 
				\cline{2-8}
				&XPIE~\cite{xia2017and_xpie} &CVPR'17 &10,000 &10,000 &[128, 300] &obj. &\makecell*[c]{covers many complex scenes with different \\numbers, sizes and positions of salient objects}	\tabularnewline  
				\cline{2-8}
				&DUTS~\cite{wang2017learning_duts} &CVPR'17 &15,572 &15,572 &[100, 500] &obj. &\makecell*[c]{from the ImageNet DET set and the SUN \\data set, very challenging scenarios}   \tabularnewline	
				\cline{2-8}
				&ILSO-1K~\cite{li2017instance_ilso1k} &CVPR'17 &1,000 &1,000 &[142, 400] &obj.\& ins. &\makecell*[c]{contains instance-level salient objects \\annotation but has boundaries roughly labeled}   \tabularnewline
				\cline{2-8}
				&SOC~\cite{fan2018salient_soc} &ECCV'18 &6,000 &6,000 &[161, 849] &obj.\& ins. &\makecell*[c]{with salient and non-salient objects from \\more than 80 common categories}   \tabularnewline
				\cline{2-8}
				&SIP~\cite{fan2020rethinking_sip} &TNNLS'20 &929 &929 &[744, 992] &obj.\& ins. &\makecell*[c]{salient person samples that cover diverse \\real-world scenes}   \tabularnewline
				\cline{2-8}
				&ILSO-2K~\cite{li2021instance_ilso2k} &CVIU'21 &2000 &2000 &[142,400] &obj.\& ins. &\makecell*[c]{most contain multiple salient object instances, \\complex background, or low color contrast. }	\tabularnewline	 
				\hline
				\hline
				&360-SOD~\cite{li2019distortion} & JSTSP’19 &500 &500 & [409, 1024] &obj. &\makecell*[c]{ERP images from five panoramic \\ video datasets with fixation groundtruth}  \tabularnewline			 	
				\cline{2-8}
				\multirow{6}{*}{\rotatebox{90}{$360^{\circ}$ ISOD}}
				&360-SSOD~\cite{ma2020stage} & TVCG’20 &1105 &1105 & [546, 1024] &obj.  & \makecell*[c]{ten categories, ERP images \\from 677 panoramic videos}   \tabularnewline
				\cline{2-8}
				&F-360iSOD~\cite{zhang2020fixation} & ICIP’20 &107 &107 & [1024, 2048] &obj.\& ins.  &\makecell*[c]{107 panoramic images,\\ 1,165 salient objects, 9 images\\ without any salient object annotations}   \tabularnewline
				\cline{2-8}
				&ODI-SOD & 2022 &6263 &6263	& [1024, 11264] &obj.  &\makecell*[c]{6263 panoramic images captured in \\real-world scenes and each image \\has pixel-wise annotation}   \tabularnewline
				\hline
			\end{tabular}
			\label{tab:all datasets}
		}
	\end{table*}

	\subsection{$360^{\circ}$ Panoramic Models}
	Large-scale 2D ISOD datasets such as DUTS~\cite{wang2017learning_duts} and XPIE~\cite{xia2017and_xpie} have extensively promoted the development of CNN-based ISOD methods~\cite{qin2019basnet,yongliangcvpr,pang2020multi,chen2020global,zhou2020interactive,wei2020label,ma2021pyramidal,liu2021visual,zhao2021complementary}.
	However, $360^{\circ}$ panoramic SOD models are minimal due to insufficient object-level pixel-wise annotation\cite{li2019distortion,ma2020stage,huang2020fanet}. 
	In \cite{li2019distortion}, a distortion adaptive module for $360^{\circ}$ ISOD is proposed to alleviate the distortion effects from the equirectangular projection by cutting the input equirectangular image into several blocks to deal with different regions with various parameters. \cite{ma2020stage} put forward a multi-stage coarse-to-fine SOD method for ODIs to handle the effects of distortions and complex scenarios using perspective images with less distortion and object-level semantical saliency ranking. Moreover, \cite{huang2020fanet} uses ERP images and much less distorted cube-map images as network input to extract and fuse features adaptively. Yet, they only focus on alleviating distortion, ignoring the effects of edge discontinuity and panoramic FoV. Similar problems also exist in other $360^{\circ}$-based tasks such as saliency prediction~{\cite{lv2020salgcn}}, object detection~{\cite{zhao2020spherical}}, panoramic semantic segmentation~{\cite{yang2021context}}, 3D room layout~{\cite{fernandez2020corners}}, and dense prediction~{\cite{tateno2018distortion}}. 
	
	For $360^{\circ}$ scenarios, the continuous panoramic view is an advantageous characteristic. Cutting a panorama into blocks or using perspective images can lose the original panoramic view and may bring more discontinuous edges, especially for changeable scale objects in complex panoramic scenes. Therefore, in this paper, we propose a $360^{\circ}$ ISOD model with the consideration of distortion, edge discontinuity and changeable scale objects in panoramic FoV.
	
	\section{Dataset}
	There are two limitations in the existing three panoramic datasets. Firstly, the most extensive dataset only contains 1105 images, which is insufficient to train a general deep network and easily leads to overfitting. Secondly, the image resolutions of the datasets are not satisfying for further research on complex $360^{\circ}$ scenarios. 
	In this section, a new large-scale dataset named ODI-SOD\footnote{The ODI-SOD dataset will be published and can be downloaded via https://github.com/iCVTEAM/ODI-SOD.git} is introduced from the aspects of dataset construction, dataset statistics and analysis. 
	
	\subsection{Dataset Construction}
	\subsubsection{Dataset Collection}
	
	The dataset ODI-SOD comprises 1,151 images collected from the Flickr website and 5,112 video frames selected from YouTube. All panoramas are in equirectangular projection format (the ratio of height and width is strictly 2:1), and the resolutions are not less than 2K. During collection, we search panoramic sources on Flickr and YouTube with different object category keywords (e.g., human, dog, building) referring to MS-COCO classes \cite{lin2014microsoft}
	to cover various real-world scenes. In this way, we collect 8,896 images and 998 videos, including different scenes (e.g., indoor, outdoor), different occasions (e.g., travel, sports), different motion patterns(e.g., moving, static), and different perspectives. Then, all videos are sampled into keyframes, and the unsatisfactory images or frames (e.g., without salient objects, low quality) are dropped out. Finally, 6,263 ERP source samples are selected for the subsequent annotation.
	
	\subsubsection{Salient Object Annotation}
	Considering most $360^{\circ}$ scenarios are complex and contain more than one object, there always exist some ambiguous objects between saliency and not saliency. It is necessary to select salient objects before time-consuming annotation. Firstly, we require five researchers to judge object saliency and select salient objects by voting. Secondly, annotation aspects manually label binary masks based on the chosen salient objects. Finally, five researchers cross-check the binary masks to ensure accurate pixel-wise object-level annotations. Some sample pairs have been shown in Fig.\ref{fig:introduction}
	
	\subsubsection{Dataset Split}
	The dataset is divided into a test set with 2,000 images and a train set with 4,263 images for deep network training. Note that all source frames from the same video are divided into the same set, train set or test set, and other source images/frames are randomly divided.

	\subsection{Dataset Statistics} 
	To explore the main characteristics of the proposed dataset and compare it with existing $360^{\circ}$ ISOD datasets, we make statistics on typical attributes of salient regions, including edge discontinuity, distortion degree and max FoV coverage.

	\subsubsection{Discontinuity of Salient Object Regions}
\begin{figure}[!t] 
		\centering
		\includegraphics[width=0.45\columnwidth]{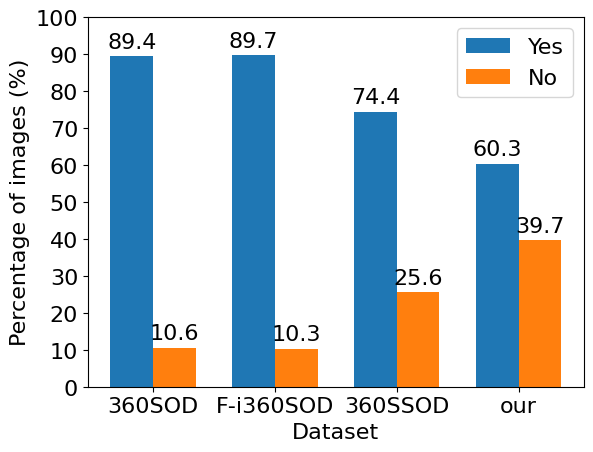}
		\caption{The percentage statistics of images with edge discontinuity (marked for Yes) and without edge discontinuity (marked for No).}
		\label{fig:edgeDis}
	\end{figure}

	Different from 2D images, the left boundary and right boundary of 360-degree images in ERP format are connected~\cite{fang2018novel}.
	For a panorama, if the central meridian crosses target salient object regions, the complete and continuous salient object regions will be divided into two discontinuous parts by the left and right boundaries of its ERP image. Here, the discontinuity of salient object regions at the boundaries is called discontinuous edge effects, which is also one of the major challenges. For the ERP images with discontinuous edge effects, it is usually more difficult to obtain complete segmentations due to the forced separation in space. Thus, it is significant to make statistics about image proportions with discontinuous edge effects. Fig.\ref{fig:edgeDis} presents the percentage of images with edge discontinuity and without edge discontinuity for the existing $360^{\circ}$ ISOD datasets. It can be seen that our dataset has a more balanced distribution and a larger number of images with edge discontinuity compared with other datasets, which is beneficial for exploring discontinuous edge effects. 
	

	\begin{figure}[!t] 
		\centering
		\subfloat[]{
			\centering
			\includegraphics[width=0.45\columnwidth]{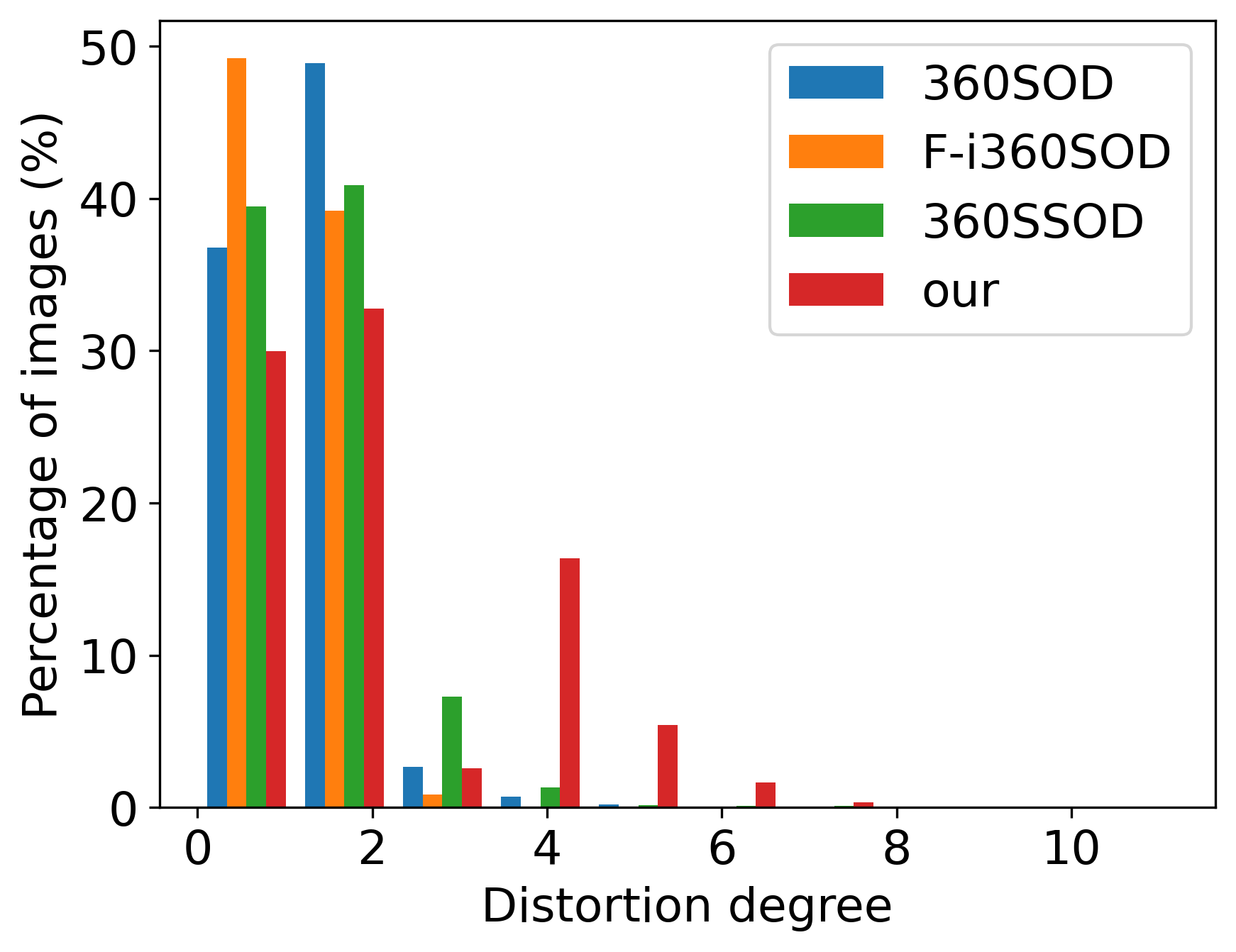} 
		}
		\subfloat[]{
			\centering
			\includegraphics[width=0.45\columnwidth]{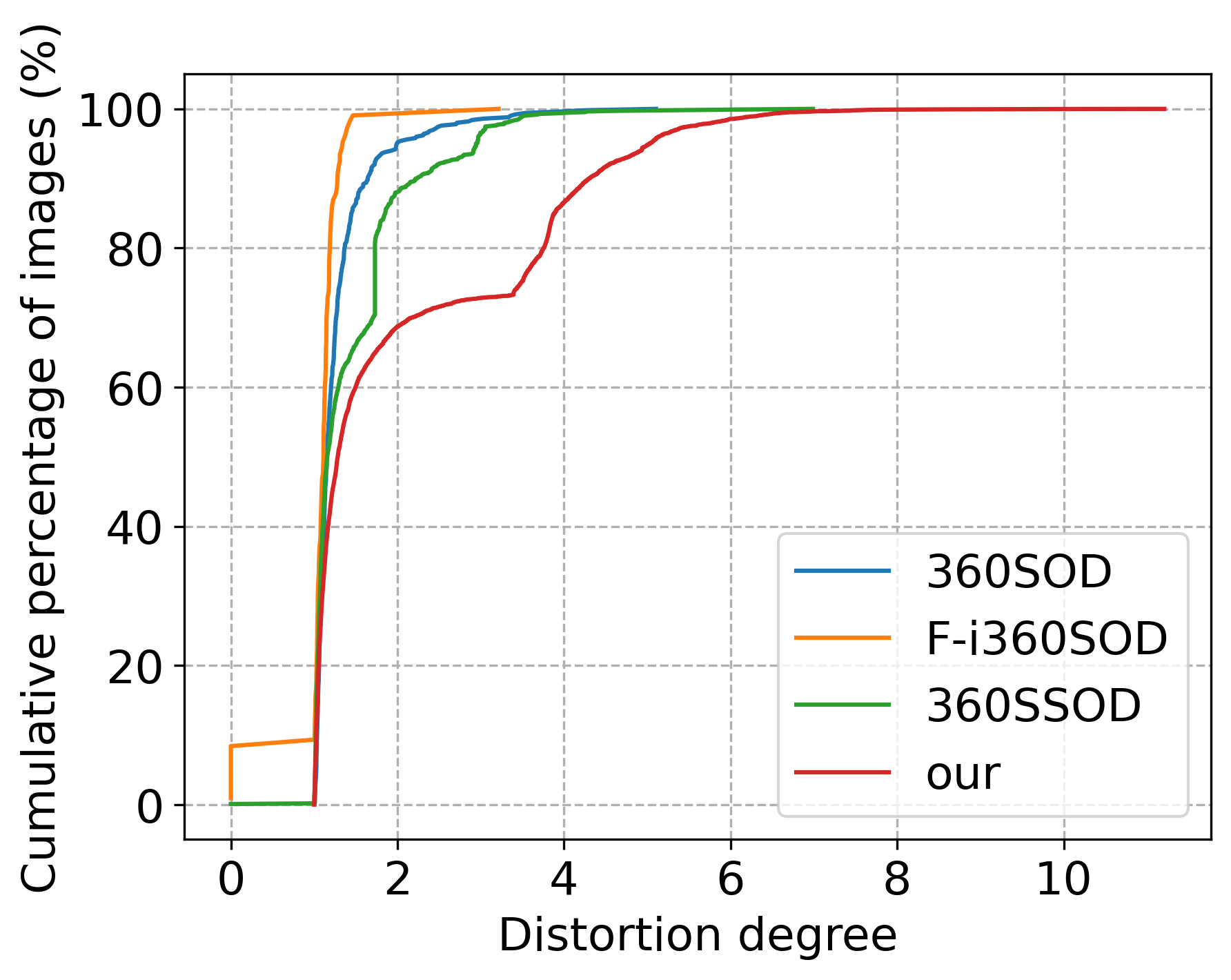} 
		}
		\caption{The statistical distribution of images with different distortion degrees. (a) The histogram distribution. (b) The cumulative distribution.}	
		\label{fig:distortionDeg}
	\end{figure}

	\subsubsection{Distortion of Salient Object Regions}
\begin{figure}[!t]
		\centering
		\includegraphics[width=0.85\columnwidth]{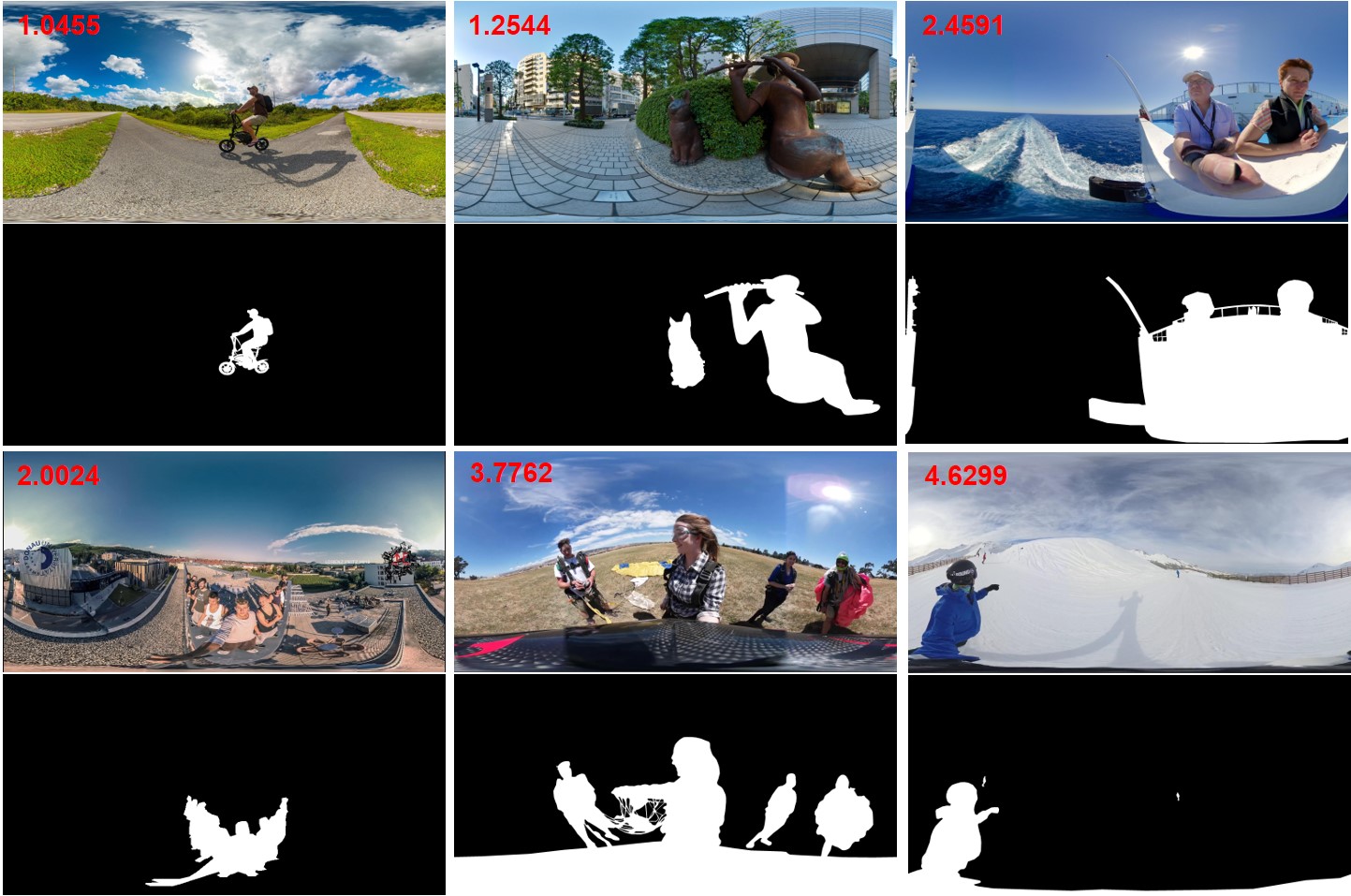}
		\caption{Examples of image and groundtruth pairs with different distortion degrees.}
		\label{fig:distortionSample}
	\end{figure}

	The distortion degrees of salient object regions usually change with their locations, reaching a maximum at the polar regions and a minimum on the equator.
	Given an ERP image $I$ and its binary groundtruth $G$ with width $w$ and height $h$, the coordinate of point $P$ can be represented as $P=(x, y)$ on the 2D pixel plane or $P=(\lambda,\varphi)$ on the sphere surface using longtitude and latitude, in which $x\in X=\{0,1,..., w-1\}, y\in Y=\{0,1,...,h-1\}$, $\lambda\in \varLambda=[-180^{\circ}, 180^{\circ}]$, $\varphi\in \varPhi=[-90^{\circ}, 90^{\circ}]$, $\varphi=E(y)$ and $E$ is the inverse projection operator. Based on $G$, we can get the salient area $a_y$ of each row on the ERP pixel plane and the corresponding area $s_{\varphi}$ on the sphere surface obtained, in which $s_{\varphi}=a_y\cdot{\cos\varphi}$.	
	To quantize the distortion, for each image, we define the distortion degree $D$ of the salient object regions as follows:
	
	\begin{equation}
		\begin{aligned}
			D=\frac{1}{n}\sum_{j=0}^{n-1}\dfrac{a_{y_j}}{s_{\varphi_j}}=\frac{1}{n}\sum_{j=0}^{n-1}\dfrac{1}{\cos{\varphi_j}}, 			  
		\end{aligned}
		\label{eq:distortion}		 
	\end{equation}
	here, $\varphi_j=E(y_j)$ and $(y_j, \varphi_j)\in Q=\{(y_j, \varphi_j) | y_j\in Y, \varphi_j\in \varPhi, s_{\varphi_j}>0\}$, and $n$ is the number of elements in set $Q$. From Eq.\ref{eq:distortion}, we can find the distortion degree of salient object regions in an ERP image mainly depends on the vertical FoV coverage and horizontal-wise area ratio of salient regions, which is a general measure of the distortion degree and has no direct relation with the number and area of salient regions. 
	Resizing all images in datasets to the same resolution and calculating their distortion values, the statistical distribution of images with different distortion degrees are counted and shown in Fig.\ref{fig:distortionDeg}. We can find that our dataset has a larger range of distortion degree distribution than other datasets. For example, in Fig.\ref{fig:distortionDeg}(b) the distortion degrees of our dataset range from 1 to about 11, while the best of others range from 0 to about 7. Besides, the proportion of images with large distortion degrees in our dataset is much larger than that in other datasets. For example, in Fig.\ref{fig:distortionDeg}(a) our dataset still has obvious distribution when the distortion degree is larger than 3. In Fig.\ref{fig:distortionDeg}(b), for other datasets the percentages of images with distortion degrees less than 3 reach more than 90\%, which means the percentages of images with distortion degrees larger than 3 are less than 10\%, while for our dataset, the percentage of images with distortion degrees larger than 3 is about 30\%.
	Moreover, in Fig.\ref{fig:distortionSample} we provide some sample pairs with different distortion degrees, presenting the reasonability of the above distortion degree formulation. 
	
	\begin{figure}[!t] 
		\centering
		\subfloat{
			\centering
			\includegraphics[width=0.85\columnwidth]{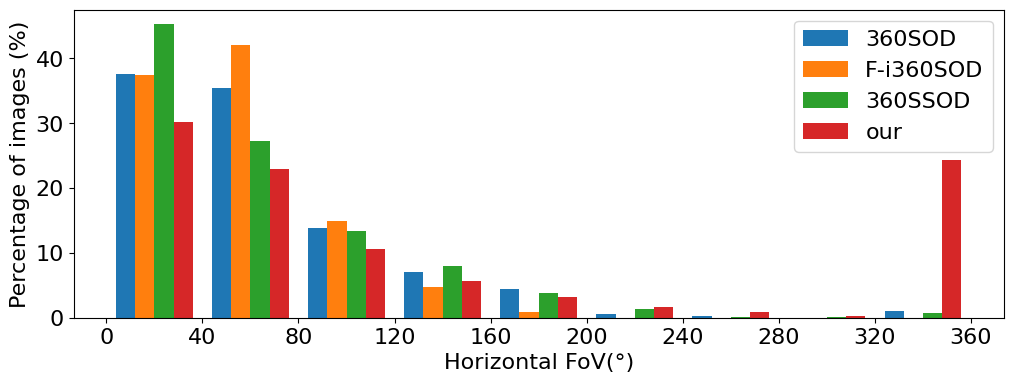}
		}
		\quad  
		\subfloat{
			\centering
			\includegraphics[width=0.85\columnwidth]{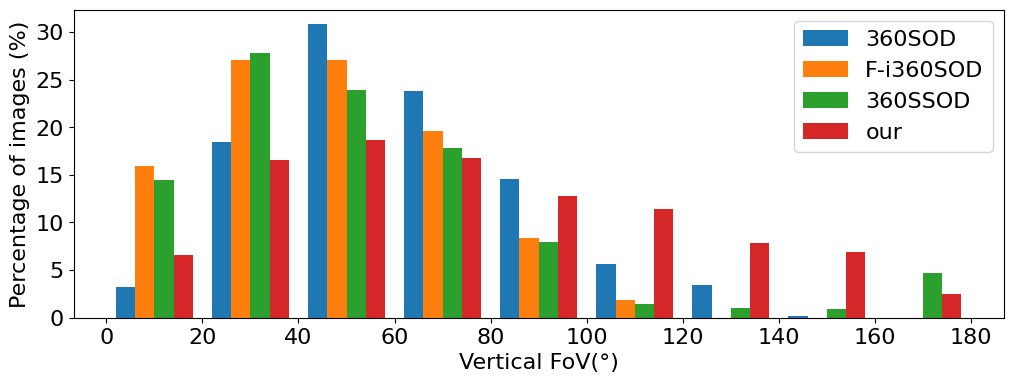}				
		}
		\caption{Horizontal and vertical FoV of salient object regions.}
		\label{fig:allFOV_hv}
	\end{figure}
	
	\subsubsection{FoV Coverage of Salient Object Regions}
	ODIs can sample the entire viewing sphere surrounding its optical center, a $360^{\circ}\times180^{\circ}$ FoV~\cite{su2017learning,zhang2020fixation}. 
	Each salient object region covers a horizontal FoV and a vertical FoV, and the covered horizontal/vertical FoV can reflect the horizontal/vertical scale of the salient object region.
	Usually, the vertical FoV coverages have more balanced distributions than the horizontal FoV coverages since salient regions are stretched more horizontally.
	To obtain holistic distributions of each dataset, calculate the max horizontal/vertical FoV coverage of salient regions in each ERP image and plot the histogram distribution in Fig.{\ref{fig:allFOV_hv}}. We find that the percentage of images decreases with the covered horizontal FoV increasing, and there are fewer images when the covered horizontal FoV is larger than $240^{\circ}$ except in our dataset. For vertical FoV, the percentage of images reaches the maximum in $[20^{\circ}, 60^{\circ}]$. The max vertical FoV coverages of most images in other datasets are smaller than $120^{\circ}$. Compared with other datasets, our dataset has more balanced and smooth distributions. It has a larger percentage of images with targets covering large FoVs, which indicates our dataset is more challenging due to the general existence of salient regions with different scales.

	\subsection{Dataset Analysis}
	From the dataset statistics, we find that discontinuous edge effects and different degrees of distortions are unavoidable due to the equirectangular projection and that different scales of salient regions are very common in complex panoramic scenes. In some cases, these characteristics can occur at the same time, which makes discontinuous edge effects, diverse distortion degrees and changeable object scales become main challenges of the $360^{\circ}$ ISOD  task. Therefore, it is necessary to design an effective model to solve above problems.

	\section{Approach}
\begin{figure*}[!t]
		\centering
		\includegraphics[width=0.9\textwidth]{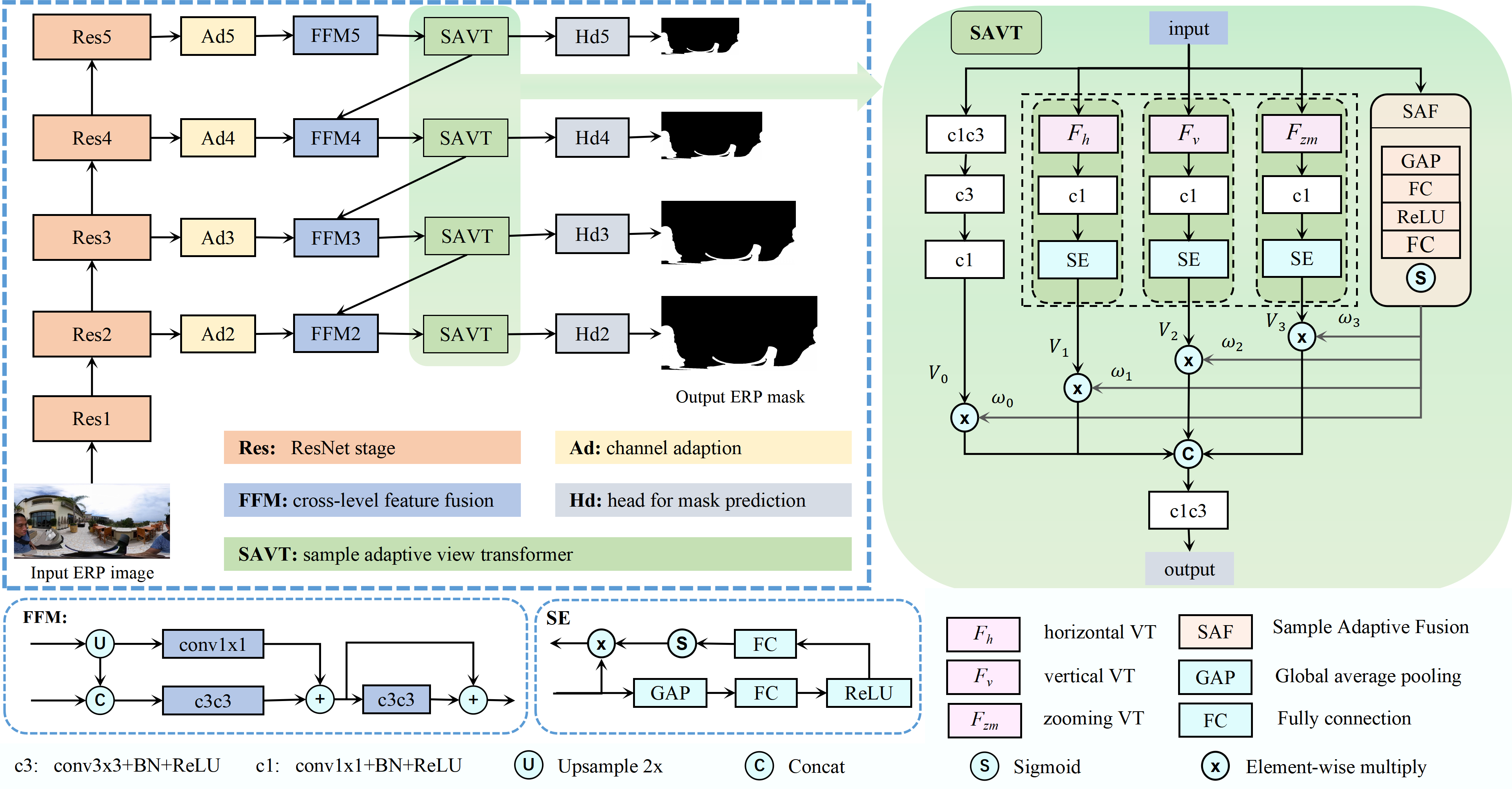} 
		\caption{The framework of our method and the proposed module SAVT with two sub-modules, View Transformer (VT) and Sample Adaptive Fusion (SAF). Specifically, VT contains three sub-branches $F_h$, $F_v$ and $F_{zm}$. $F_h$ means horizontal view transformer, $F_v$ means vertical view transformer and $F_{zm}$ means zooming view transformer. 
		}
		\label{framework}
	\end{figure*}

	\noindent To overcome above problems, we present our overall approach as shown in Fig.\ref{framework}, which consists of the encoder, decoder and the proposed Sample Adaptive View Transformer (SAVT) module that has two sub-modules View Transformer (VT) and Sample Adaptive Fusion (SAF).
	To better understand the two modules, we first introduce basic concepts in Sec.~\ref{sec:Pre}. Then, we describe the overall framework of our method in Sec.~\ref{sec:method_overall}. Subsequently, Sec.~\ref{sec:DT} illustrate the proposed SAVT in detail.

	\begin{figure}[!t]
		\centering
		\includegraphics[width=0.8\columnwidth]{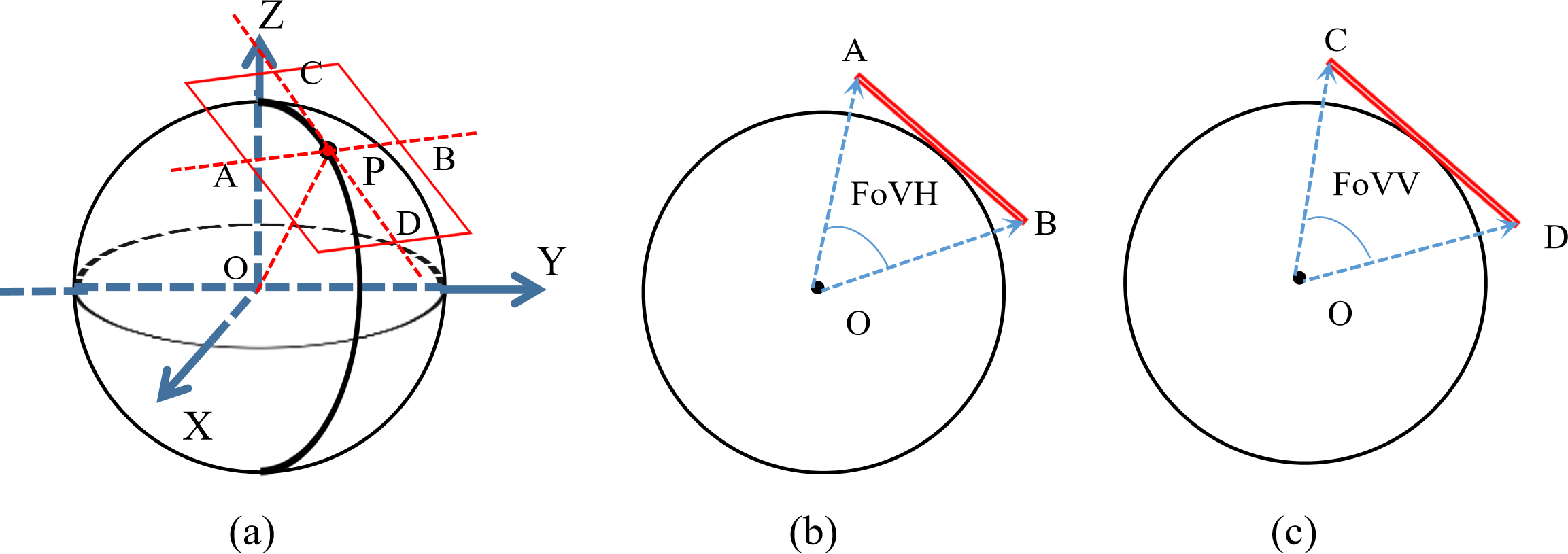} 
		\caption{A diagram of the viewport. P is the tangent point of the sphere and viewport, i.e., the viewpoint. Points A, B, C, and D are the center points of viewport edges. FoVH and FoVV are the horizontal and vertical fields of view, respectively.}
		\label{VP}
	\end{figure}
	
	\subsection{Preliminary}\label{sec:Pre}
	In this part, we briefly introduce the process and characteristics of the equirectangular projection and explain the viewport and viewpoint used in panoramas and Möbius transformations.

	\subsubsection{Viewport}\label{sec:Viewport}

	As shown in Fig.\ref{VP}, in panoramic vision, when looking at the point $P(\lambda,\varphi)$ from the sphere center $O$ with the horizontal and vertical field of view FoVH and FoVV, respectively, we can see a region $R$ of the sphere surface, and $P$ is the center of view (i.e., viewpoint). The points on the sphere region $R$ can be projected to a rectangular plane tangent to the sphere surface at point $P$ by gnomonic projection. 
	The tangent plane is defined as the viewport as \cite{de2016geometry, yu2015framework} do. The distance from $O$ to $P$ is called view distance in the study.
	
	\subsubsection{Möbius Transformation}	
	Möbius transformations
	are one-to-one, onto and conformal (angle preserving) maps of the so-called extended complex plane \cite{olsen2010geometry}. 
	The extended complex plane is given by $\mathbb{C}_{\infty}=\mathbb{C} \cup {\infty}$. 
	A Möbius transformation $f:\mathbb{C}_{\infty}\rightarrow \mathbb{C}_{\infty}$ is a map
	\begin{equation}
		\label{eq:Moebius}
		f(z)=\dfrac{az+b}{cz+d}, \quad a,b,c,d\in \mathbb{C}, ad-bc\neq0,
	\end{equation}
	where a,b,c and d are constant complex numbers satisfying $ad-bc\neq0$.
	
	The Riemann sphere is a model of $\mathbb{C}_{\infty}$, which is homeomorphic to the two-dimensional sphere\cite{blanchard1984complex} 
	\begin{equation}
		S^2=\{(x_s,y_s,z_s)\in \mathbb{R}^3 | x_s^2+y_s^2+z_s^2=1\}.
	\end{equation}
	We identify the complex plane $\mathbb{C}$ with the equitorial plane $x_3=0$, and set the North pole $N=(0,0,1)$. If the line from $N$ to $P$ intersects the complex plane in exactly one point $z\in\mathbb{C}$, then the map $SP:S^2\setminus N \rightarrow \mathbb{C}$ which assigns a point $P\in S^2$ to the point $z\in\mathbb{C}$ is called the stereographic projection~\cite{olsen2010geometry}, which 
	is a bridge between $R^3$ and $\mathbb{C}_{\infty}$. For $P=(x_s,y_s,z_s)\in S^2 \setminus N$ and $z=x+iy\in \mathbb{C}$, $SP$ is given by 
	\begin{equation}
		SP: (x_s,y_s,z_s)\rightarrow \dfrac{x_s}{1-z_s}+i\dfrac{y_s}{1-z_s}.
	\end{equation}

	For ODIs, ERP is just one of the projection formats. The vanilla representation is the sphere surface representation, also called visible sphere, fully representing the original field of view with $360^{\circ}$ longitude by $180^{\circ}$ latitude. Therefore, Möbius transformations can be applied to ODIs, which is vital for the proposed method.

	\subsection{Overall Framework}\label{sec:method_overall}

	To take advantage of existing mature 2D CNNs, 
	we take into account the classic U-shape structure and add the proposed tailored delicate SAVT module aiming at ODIs. The overall framework is shown in Fig.~\ref{framework}. The input and output are both in the ERP format.
	
	For the encoder, the backbone network uses ResNet-50~\cite{he2016deep} removed the last global pooling and fully connected layers for the pixel-level prediction. 
	For the decoder, the output features of the encoder pass through the channel adaption modules and feature fusion modules FFM, and then transmit to SAVT. Each FFM connecting with $\{Res2, Res3, Res4\}$ fuses the features of the current stage and adjacent higher stage into enhanced features for SAVT in the current stage. The FFM in stage 5 ignores the upsample interpolation and concat operations at the entrance. Each SAVT connects with a mask head consisting of a convolution layer with kernel $3*3$ as the channel compression layer and an upsampling interpolation operation. All mask heads are used for side-output supervision in the training stage.
	The progressive strategy from coarse to fine is beneficial for SOD tasks. 
	
	The proposed SAVT contains two parallel sub-modules, View Transformer (VT) and Sample Adaptive Fusion (SAF). VT has three branches $F_h, F_v, F_{zm}$ corresponding to different transformations to simulate the human observing process of changing viewpoints or view distances, and SAF is used to adjust the output values of other parallel branches. Next, we present it in detail.

	\subsection{Sample Adaptive View Transformer}\label{sec:DT}

	\subsubsection{View Transformer Types}
	
	No matter the ODIs are displayed in the desktop setting or head-mounted VR, what can be seen is very limited at each moment. We have to change our viewpoint or adjust view distance to obtain more information. Inspired by this, for ERP image processing, we introduce rotation and zooming, the two kinds of transformation, to simulate the observing process of looking left and right (branch $F_h$), up and down (branch $F_v$), far and near (branch $F_{zm}$).

	\subsubsection{View Transformer Formulation}	

	An ODI represented as the Riemann sphere can use different Möbius transformations, making the panoramic scene keep continuous in panoramic view after transformation.

	About rotation, a map $f: \mathbb{C}_{\infty} \rightarrow \mathbb{C}_{\infty}$ is called a rotation of $\mathbb{C}_{\infty}$ if the map $SP^{-1} \circ f \circ SP:S^2 \rightarrow S^2$ is a rotation~\cite{olsen2010geometry}. 
	Möbius transformations represent rotations if and only if $c=-\overline{b}, d=\overline{a}$, and $ad-bc=|a|^2+|b|^2=1$, i.e.,
	\begin{equation}
		\label{eq:MoebiusRotation}
		f(z)=\dfrac{az+b}{-\overline{b}z+\overline{a}}, \quad a,b,c,d\in \mathbb{C}, ad-bc=1.
	\end{equation}
	For convenience, under a rotation of an angle $\theta \in [0,2\pi]$ about the axis passing through the origin in the direction along the vector $\textbf{L}=(l,m,n)$, based on the formulas of stereographic projection and Riemann sphere~\cite{Mobius_2017}, the complex number $a, b$ in $f(z)$ can be derived as follows:
	\begin{align}
		a &= cos(\theta/2) + i\cdot n \cdot sin(\theta/2), \\
		b &= (m-i\cdot l)sin(\theta/2).
	\end{align}
	If $\textbf{L}=(0,0,1)$, then $a=cos(\theta/2)+i\cdot sin(\theta/2), b=0$, 
	then there is 
	\begin{equation}
		f(z)=\frac{az}{\overline{a}}=e^{i\theta}z, 
	\end{equation}
	namely the canonic \textit{elliptic} Möbius transformation, which can simulate looking left or right. Similarly, if $\textbf{L}=(0,1,0)$, then it can simulate looking up or down.
	
	About zooming, we set $c=0$ and simplify Eq.\ref{eq:Moebius} as follows:

	\begin{equation}
		f(z) = az,  \quad a\in \mathbb{C}.
	\end{equation}
	To facilitate, write $a=\rho e^{i\theta}$. When $\theta=0, \rho<1, f(z)$ is an origin-centered contraction, and when $\theta=0, \rho>1, f(z)$ is an origin-centered expansion. The Möbius transformations with contraction or expansion are called \textit{hyperbolic}. We can change the zooming center by rotating the target center to the origin and inversely rotating it back after zooming.

	Before applying to the feature space, we take the transform process on an ERP image as an example for a simplified description and better visualization of the geometric transformation. 
	For each point $P_e(u,v)$ on the ERP pixel plane, there is a corresponding point $P_s(\lambda,\varphi)$ ($\lambda$ is longtitude, and $\varphi$ is latitude) on the sphere surface. The coordinate of $P_s$ in $\mathbb{R}^3$ is $P(x_s,y_s,z_s)$. Usually, the $P_s(0,0)$ is projected to the center of equirectangular image. The transform relationship can be represented as
	\begin{equation} 
		P_e(u,v)=T(P(x,y,z)),
	\end{equation}
	\begin{equation}
		P(x,y,z)=T^{-1}(P_e(u,v)),
	\end{equation}
	where $T(\cdotp)$ is the transform function from the sphere surface to the ERP pixel plane, and $T^{-1}(\cdotp)$ is its inverse transform function. 
	
\begin{figure}[!t]
		\centering
		\includegraphics[width=0.9\columnwidth]{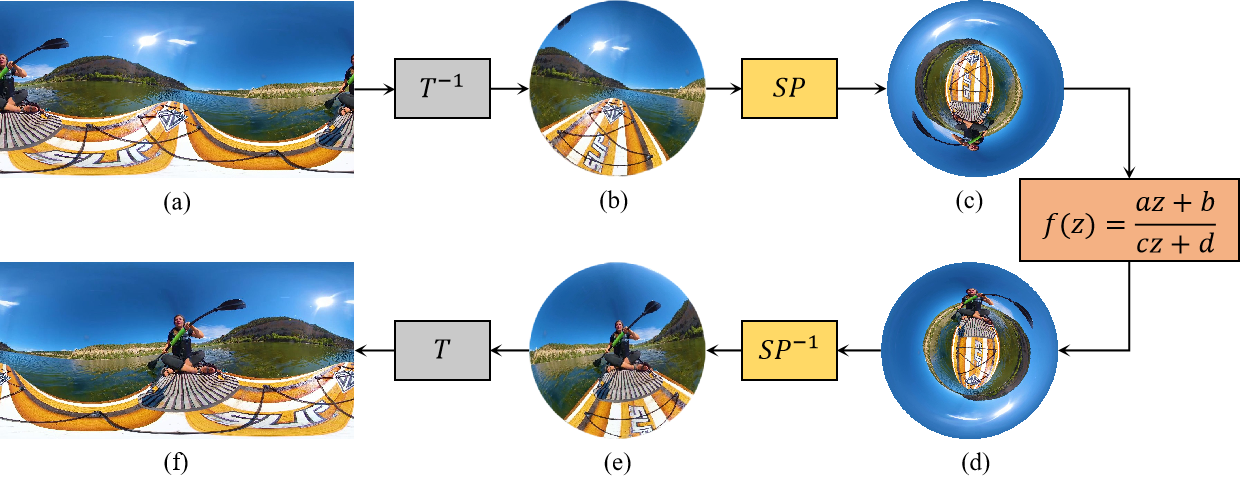} 
		\caption{Transform steps in View Transformer.}
		\label{fig:DT_steps}
	\end{figure}

\begin{figure}[!t]
		\centering
		\includegraphics[width=0.95\columnwidth]{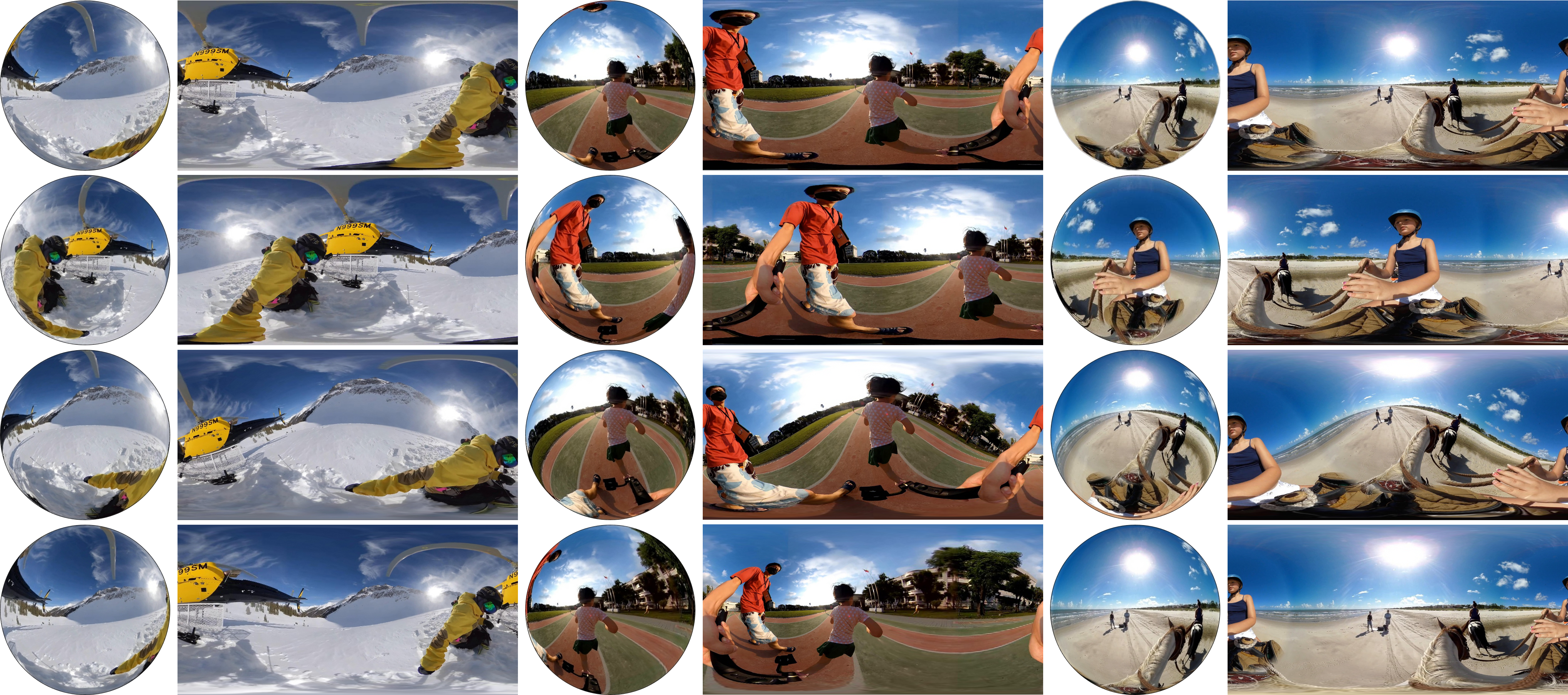}
		\caption{Some transformation examples by View Transformer. Each row from top to bottom corresponds to the original images, horizontal transformation, vertical transformation, and zooming transformation. From left to right, the horizontal transformation parameters are $\theta=150^{\circ}, 100^{\circ}$ and $180^{\circ}$; the vertical transformation parameters are all $\theta=30^{\circ}$; the zooming transformation parameters are $\rho=1.5 (O=(0,1,0)), \rho=0.5 (O=(0,1,0))$ and $\rho=1.5 (O=(-1,0,0))$.}
		\label{fig:example_VT}
	\end{figure}

	As shown in Fig.\ref{fig:DT_steps}, for a given ERP image, the whole process is divided into five steps: 1) first back-project the ERP pixel plane to the sphere surface; 2) through stereographic projection $SP$, get its representation in the extended complex plane $\mathbb{C}_{\infty}$, 3) make Möbius transformations in $\mathbb{C}_{\infty}$; 4) back-project to the Riemann sphere after transformations; 5) project to the ERP pixel plane from the sphere surface. The simplified formula is as follows:
	\begin{equation}
		\begin{split}
			P_e^{'} &= T(SP^{-1}(f(SP(T^{-1}(P_e))))).\\
			&= F(P_e).
		\end{split}
	\end{equation}\label{eq:transPe}
	Here, $P_e^{'}$ is the point on the ERP image after view transformation. The whole process of transformation is reversible, represented as $F^{-1}(\cdotp)$.
	By doing this, for an ERP image, we can get the transformation images under different views, and the panoramic views are kept simultaneously.
	Fig.~\ref{fig:example_VT} shows some transformation examples of images. It can be seen that transformations keep the complete panoramic view and obtain appearances under diverse views.
	 
	\subsubsection{Refine the Feature Space}
	\begin{figure}[!t]
		\centering
		\includegraphics[width=0.98\columnwidth]{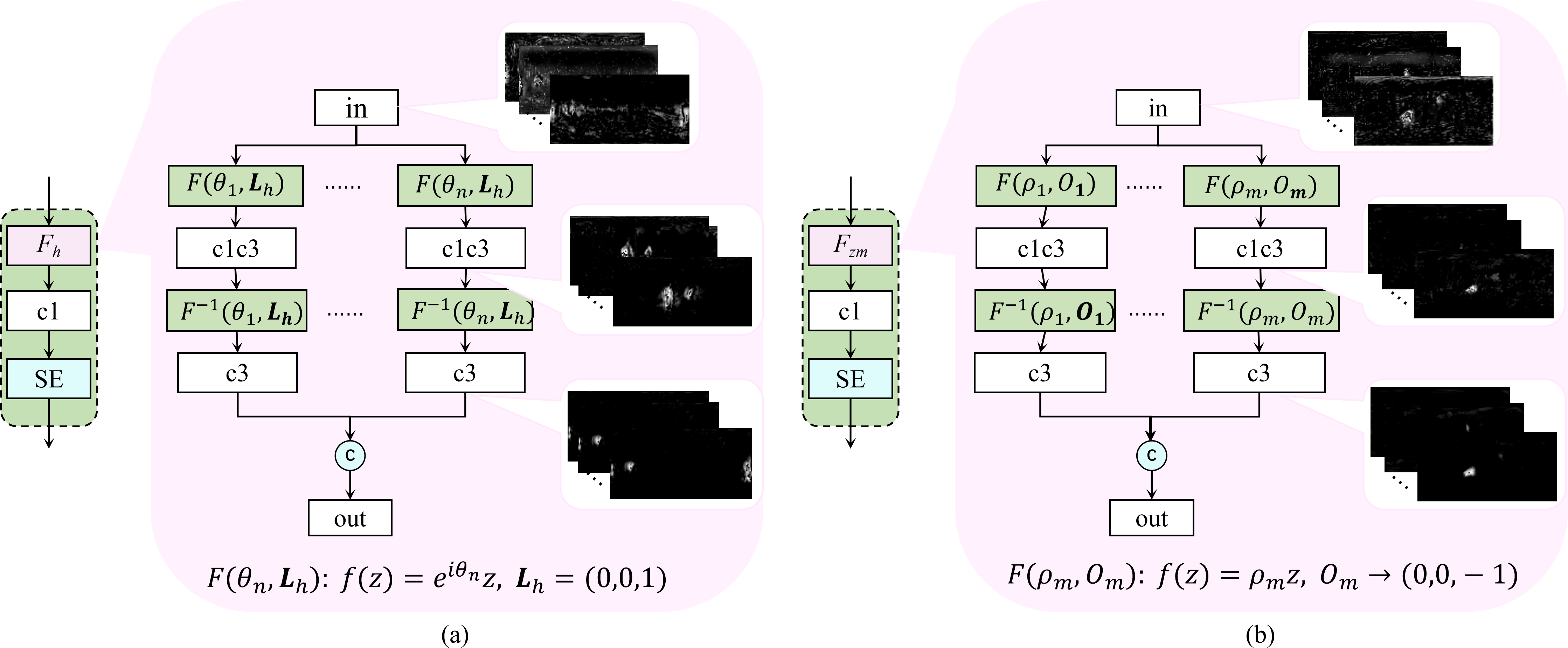} 
		\caption{View Transformer branches. (a) Horizontal rotation branch with $F_h$. (b) Zooming branch with $F_{zm}$. In (a), $\theta_{i},i\in[1,n]$ is the rotation degree and $L_h$ is the rotation axis. In (b), $\rho_{i},i\in[1,m]$ is the zooming factor and $O_m$ is zooming center.}
		\label{fig:DT}
	\end{figure}	
	
	In the study, we use the transforming process to refine the feature space. Then, each $P_e$ or $P_e^{'}$ corresponds to a feature position before or after transformation. 
	The feature value $v_p$ at the feature point $p=(i,j,k)$ is mapped to the target position $p'=(i',j',k)$ through transformation, in which $i,i'\in[0, h-1], j,j'\in [0,w-1]$, $k\in[0,c-1]$, $h$ is the height of the input feature, $w$ is the width and $c$ is the channels. Theoretically, in the pure geometric transformation, the feature value is unchanged. The geometric transformations contribute to rich feature appearances under different views, which is view-aware. Thus, by feat of the transformed features, we can learn more appropriate features and fuse them after inverse transformation.
	
	Fig.\ref{fig:DT} shows more details about branches of VT, in which the vertical branch is not contained because it is similar to the horizontal branch except for the vector $L_v=(0,1,0)$. $F_h, F_v, F_{zm}$ contain different numbers of sub-branches, and each sub-branch corresponds to different transform parameters. For $F_h$, $L_h$ is fixed and the degree $\theta_{i}, i\in[1,n]$ is the horizontal rotation degree of looking left or right. For $F_v$, $L_v$ is fixed and the degree parameter is the vertical rotation degree of looking up or down. For $F_{zm}$, $O_m$ is like our viewpoint and $\rho_{i},i\in[1,m]$ controls looking near or far. 	
	
	\subsubsection{Sample Adaptive Fusion}\label{sec:SAF}
	
	To make better use of these transformation features, we perform an adaptive fusion of these features to adapt to different samples. This fusion process is expected to be simple and efficient.
	Here, we use a SENet block \cite{hu2018squeeze} to realize it by learning an adaptive weight for each type of transformation branch and the original learning branch (see Fig~\ref{framework}). Then fuse the weighted features by a concat operation in the channel dimension, as follows:
	\begin{equation}
		V_f = Concat(\omega_{k}\cdot V_{k}), {\quad k=0,1,2,3,}
	\end{equation}
	where $V_0$ corresponds to the original feature learning branch without any geometric transformation, $V_1$ and $V_2$ correspond to the two rotation branches, $V_3$ corresponds to the zooming branch, and $\omega_k$ is the function of the original input feature which depends on the input sample. 
	Thus, we called the process Sample Adaptive Fusion. 
	Through SAF, the transformed features can be adaptively fused and better represent the current sample.
	
\begin{figure}[!ht]
		\centering
		\includegraphics[width=0.95\columnwidth]{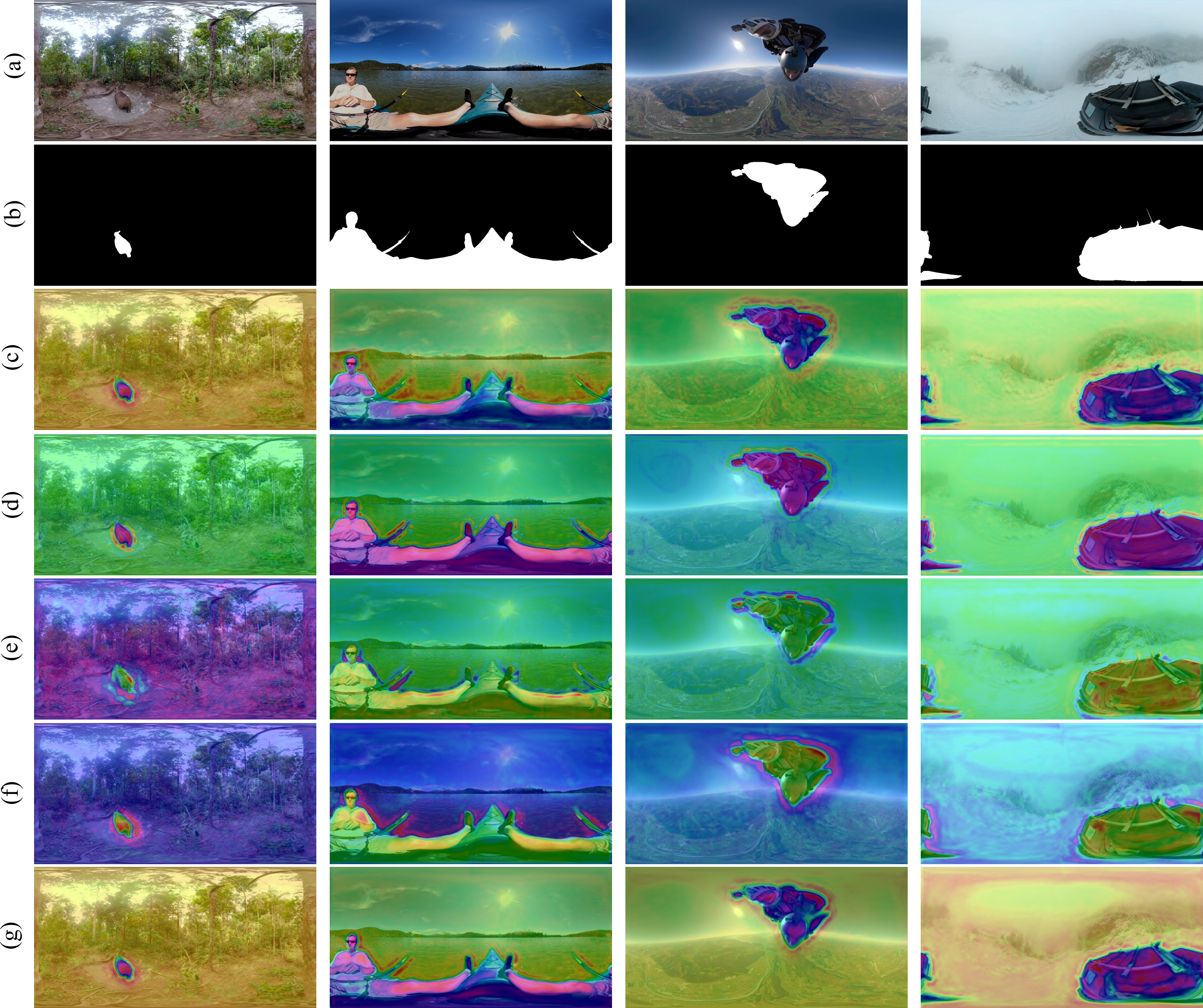} 
		\caption{Visual examples of gradient CAMs of the SAVT module. (a) Input images. (b) Groundtruth maps. (c) CAMs based on input features of SAVT. (d) CAMs based on output features of SAVT. (e) CAMs based on output features of the horizontal rotation branch. (f) CAMs based on output features of the vertical rotation branch. (g) CAMs based on output features of the zooming branch.}
		\label{fig:feature_SADT}
	\end{figure}

	Fig.\ref{fig:feature_SADT} shows the gradient class activation maps (CAMs)~\cite{selvaraju2017grad} of some representative samples with objects that are obviously distorted or with discontinuous edge effects or on large or small scales. 
	From Fig.\ref{fig:feature_SADT} we find that the output features (the forth row) of SAVT are much better than its input (the third row), which indicates SAVT is effective. Specifically, there always exists at least one transform branch outputting better features in some regions for different samples, which suggests that VT offers diverse and helpful candidate features.
	Moreover, the output features of different branches are integrated better by SAF adjusting weights, which shows VT and SAF are combined effectively. 
	
	\section{Experiments}
	In the section, we first benchmark current state-of-the-art 2D-based SOD methods and $360^{\circ}$-based SOD methods on the proposed dataset ODI-SOD. Then we choose representative methods to train on our dataset and compare with the proposed method. Furthermore, we verify and analyze the effectiveness of the proposed method in the ablation study.
	
	\subsection{Experimental Settings}
	
	\subsubsection{Dataset and Evaluation Metrics}
	We conduct experiments on the proposed dataset ODI-SOD, which contains 4,263 training images and 2,000 testing images.	
	To quantitatively evaluate the performance of methods, we utilize the common metrics for SOD, namely the mean absolute error (MAE), F-meansure ($F_\beta$), weighted F-measure($wF_\beta$)\cite{margolin2014evaluate}, max F-measure ($maxF$), S-measure ($S_m$)\cite{fan2017structure}, E-measure ($E_m$)\cite{fan2018enhanced, fan2021cognitive}.
	F-measure indicates the trade-off result between precision and recall and 
	here we set $\beta^2=0.3$ to emphasize more precision than recall as in \cite{achanta2009frequency}.

	\subsubsection{Implementation Details}
	In the training stage, we use the pre-trained ResNet-50 model~\cite{he2016deep} to initialize the parameters of the feature encoder and use a standard stochastic gradient descent algorithm to train the whole network end-to-end with the cross-entropy loss and IoU loss. In our network encoder, the initial learning rate is set to 0.05 with a weight decay of 0.0005 and momentum of 0.9. For the rest layers, the learning rates are ten times the encoder. We train the proposed method with a mini-batch of size 16 about 64 epochs by a single GTX 3080 GPU.
	In the testing stage, only the output of $Hd2$ in Fig.\ref{framework} is used for the final prediction result. In both training and testing, the input images are resized to $512\times 256$ resolution for comparison with other $360^{\circ}$ SOD methods.

	\begin{table*}[!htp]
		\centering{
			\caption{Benchmarking results of the SOTA methods on the ODI-SOD test set before training by ODI-SOD train set.}
			\label{tab:benchmark}
			\begin{tabular}{l|l|l|l|l|cccccc}
				\hline
				\textbf{Methods}       & \textbf{Year} & \textbf{Type}               & \textbf{Backbone} & \textbf{Train set} & MAE$\downarrow$ & $F_\beta$$\uparrow$ & $wF_\beta$$\uparrow$ & $S_m$$\uparrow$ & $E_m$$\uparrow$ & $maxF$$\uparrow$ \\ \hline
				GCPANet~\cite{chen2020global}              & 2020 AAAI     & 2D SOD                        & ResNet 50         & DUTS-TR            & 0.128           & 0.460               & 0.428                & 0.647           & 0.669           & 0.568            \\ \hline
				MINet-R~\cite{pang2020multi}              & 2020 CVPR     & 2D SOD                        & ResNet 50         & DUTS-TR            & 0.123           & 0.435               & 0.399                & 0.624           & 0.664           & 0.528            \\ \hline
				ITSD~\cite{Zhou_2020_CVPR}              & 2020 CVPR       & 2D SOD                           & ResNet 50         & DUTS-TR            & 0.137           & 0.450                & 0.427               & 0.635           & 0.655           & 0.538            \\ \hline
				F3Net~\cite{wei2020f3net}              & 2020 AAAI     & 2D SOD                          & ResNet 50         & DUTS-TR            & 0.133           & 0.423               & 0.387                & 0.615           & 0.655           & 0.519            \\ \hline
				DFI~\cite{liu2020dynamic}              & 2020 TIP         & 2D SOD                            & ResNet 50         & DUTS-TR            & 0.108           & 0.460                & \color{blue}{0.430} &\color{blue}{0.654} & 0.674           & 0.570             \\ \hline
				PFSNet~\cite{ma122021pyramidal}              & 2021 AAAI     & 2D SOD                         & ResNet 50         & DUTS-TR            & 0.141           & 0.421               & 0.388                & 0.609           & 0.649           & 0.514            \\ \hline
				CTDNet~\cite{zhao2021complementary}              & 2021 MM       & 2D SOD                         & ResNet 50         & DUTS-TR            & 0.138           & 0.423               & 0.389                & 0.610           & 0.662           & 0.509            \\ \hline
				VST~\cite{liu2021visual}              & 2021 ICCV        & 2D SOD                            & T2T-ViT           & DUTS-TR            & 0.135           & 0.428               & 0.402                & 0.621           & 0.656           & 0.518            \\ \hline
				PAKRN~\cite{xu2021locate}               & 2021 AAAI     & 2D SOD                          & ResNet 50         & DUTS-TR            & \color{blue}{0.092} & \color{red}{0.556} & \color{red}{0.518} & \color{green}{0.694} & \color{green}{0.729}  & \color{green}{0.642}    \\ \hline
				DCN~\cite{wu2021decomposition}               & 2021 TIP      & 2D SOD                            & ResNet 50         & DUTS-TR            & 0.125           & 0.417               & 0.383                & 0.613           & 0.648           & 0.514            \\ \hline
				SOD100K~\cite{cheng2021highly}              & 2021 TPAMI    & 2D SOD                        & ResNet 50         & DUTS-TR            & 0.198           & 0.288               & 0.245                & 0.544           & 0.543           & 0.401            \\ \hline
				PSGLoss~\cite{yang2021progressive}              & 2021 TIP      & 2D SOD                        & ResNet 50         & DUTS-TR            & 0.116           & 0.439               & 0.392                & 0.616           & 0.675           & 0.521            \\ \hline
				SCASOD~\cite{Siris_2021_ICCV}              & 2021 ICCV     & 2D SOD						 & ResNet 50         & DUTS-TR            & \color{green}{0.083} & 0.455          & 0.391                & 0.625           & 0.582           & 0.475            \\ \hline
				FastSaliency~\cite{liu2021samnet}              & 2021 TIP      & 2D SOD                   & ResNet 50         & DUTS-TR            & 0.185           & 0.319               & 0.287                & 0.557           & 0.575           & 0.429            \\ \hline
				PurNet~\cite{li2021salient}              & 2021 TIP      & 2D SOD                         & ResNet 50         & DUTS-TR            & 0.119           & 0.436               & 0.400                & 0.622           & 0.678           & 0.535            \\ \hline
				PoolNet~\cite{Liu2019PoolSal}              & 2022 TPAMI    & 2D SOD                        & ResNet 50         & DUTS-TR            & 0.101           & 0.466               & 0.419                & 0.647           & 0.687           & \color{blue}{0.576}  \\ \hline
				RCSB~\cite{Ke_2022_WACV}              & 2022 WACV     & 2D SOD                           & ResNet 50         & DUTS-TR            & 0.108           & \color{blue}{0.490} & 0.427                & 0.630           & \color{blue}{0.692}  & 0.561            \\ \hline
				ZoomNet~\cite{ZoomNet-CVPR2022}              & 2022 CVPR     & 2D SOD                        & ResNet 50         & DUTS-TR            & 0.120           & 0.465               & 0.429                & 0.644           & 0.670           & 0.558            \\ \hline
				TRACER~\cite{lee2021tracer}              & 2022 AAAI     & 2D SOD                         & ResNet 50         & DUTS-TR            & 0.099           & 0.460               & 0.418                & 0.630           & 0.691           & 0.530             \\ \hline \hline
				DDS~\cite{li2019distortion} & 2019 JSTSP    & $360^{\circ}$ SOD                            & ResNet 50         & 360-SOD            &\color{red}{0.070} & \color{green}{0.553} & \color{green}{0.493} & \color{red}{0.694} & \color{red}{0.751}  & \color{red}{0.648} \\ \hline
			\end{tabular}			
			
		}	
	\end{table*}

	\begin{figure*}[!t] 
		\centering
		\subfloat{
			\begin{minipage}[t]{0.85\linewidth}
				\centering
				\includegraphics[width=0.98\columnwidth]{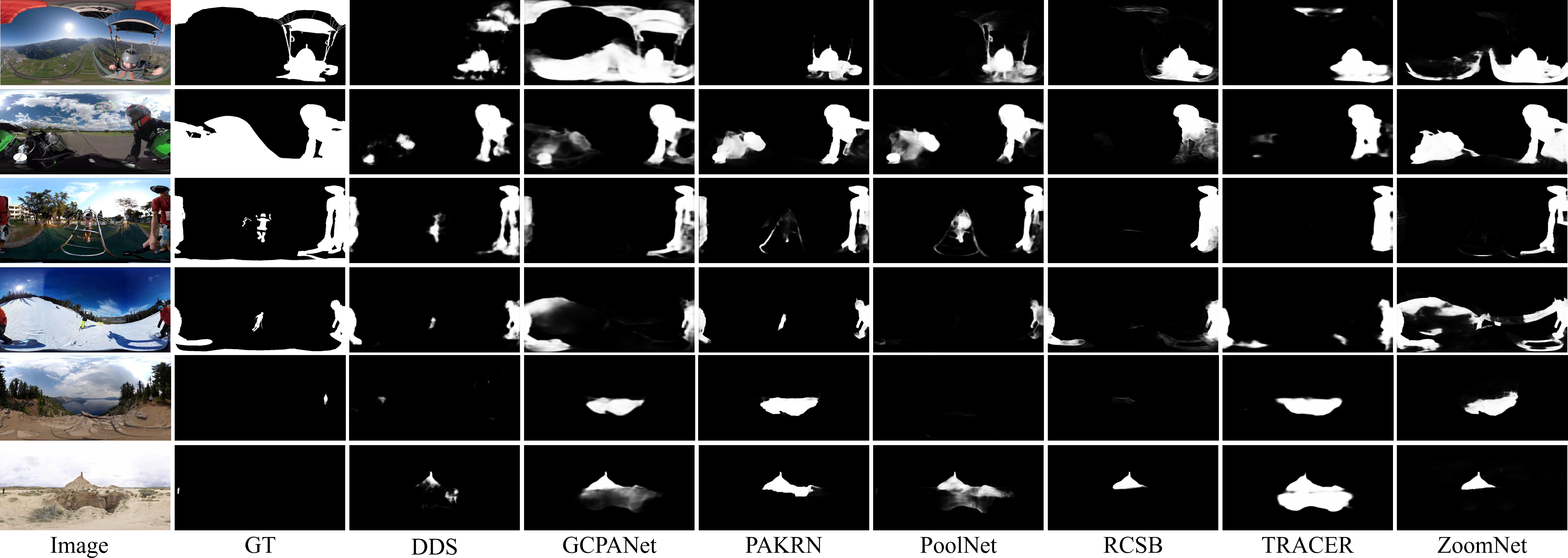}
			\end{minipage}
		}
		\caption{Visual testing examples of representative state-of-the-art algorithms before finetuning on the ODI-SOD train set. }
		\label{fig:benchmark_wo_finetune}
	\end{figure*}

	\subsection{Benchmarking Results}

	To verify the challenges of the proposed dataset ODI-SOD, in Tab.\ref{tab:benchmark} we list the performances of 20 state-of-the-art (SOTA) 2D SOD and $360^{\circ}$ SOD methods on our test set without finetuning on our train set. The methods include GCPANet~\cite{chen2020global}, MINet-R~\cite{pang2020multi}, ITSD~\cite{Zhou_2020_CVPR}, F3Net~\cite{wei2020f3net}, DFI~\cite{liu2020dynamic}, PFSNet~\cite{ma122021pyramidal}, CTDNet~\cite{zhao2021complementary}, VST~\cite{liu2021visual}, PAKRN~\cite{xu2021locate}, DCN~\cite{wu2021decomposition}, SOD100K~\cite{cheng2021highly}, PSGLoss~\cite{yang2021progressive}, SCASOD~\cite{Siris_2021_ICCV}, FastSaliency~\cite{liu2021samnet}, PurNet~\cite{li2021salient}, PoolNet~\cite{Liu2019PoolSal}, RCSB~\cite{Ke_2022_WACV}, ZoomNet~\cite{ZoomNet-CVPR2022}, TRACER~\cite{lee2021tracer} and DDS~\cite{li2019distortion}. 
	From Tab.\ref{tab:benchmark} we find that all listed methods perform not well on the ODI-SOD test set including the 360$^{\circ}$-based method DDS~\cite{li2019distortion} trained by the 360-SOD train set, which suggests that currently available models have poor generalization ability over the proposed dataset. It comes down to two reasons. Firstly, a gap exactly exists between 2D SOD datasets and 360$^{\circ}$ omnidirectional SOD datasets, which makes the outstanding 2D methods have sharp drops in performance. Secondly, the proposed dataset is very challenging and beyond the cognitive capabilities of the existing datasets and models.

	For further comprehensive analysis, some testing results of representative superior SOTA algorithms in Tab.\ref{tab:benchmark} are shown in Fig.\ref{fig:benchmark_wo_finetune}. 
	From the prediction maps, we find that the less distorted or evident salient target regions can be handled by most of the methods. 
	In contrast, the severely distorted regions can easily lead to segmentation failure due to their apparent differences from existing knowledge and perception. The target objects with discontinuous edge effects and the small-scale or large-scale objects are also difficult to be completely segmented out. 
	It illustrates that severe distortions, discontinuous edge effects and changeable scales are three major challenges in the proposed $360^{\circ}$ dataset.

	\subsection{Comparison with State-of-the-arts}

	\begin{table*}[!htp]
		\centering{
			\caption{Performance comparison of our method, $360^{\circ}$-based method and the top-5 2D-based SOTA methods on the ODI-SOD test set after finetuning on the ODI-SOD train set.}
			\label{tab:PerformanceComp}
			\begin{tabular}{l|l|l|c|c|cccccc}
				\hline
				\textbf{Methods}       & \textbf{Year} & \textbf{Type}              & \textbf{Params (M)} & \textbf{MACs (G)} & MAE$\downarrow$ & $F_\beta$$\uparrow$ & $wF_\beta$$\uparrow$ & $S_m$$\uparrow$ & $E_m$$\uparrow$ & $maxF$$\uparrow$ \\ \hline
				PAKRN~\cite{xu2021locate}              & 2021 AAAI     & 2D SOD                         & 141.06         & 228.85           & 0.106           & 0.408               & 0.410                & 0.632           & 0.611           & 0.727            \\ \hline
				PoolNet~\cite{Liu2019PoolSal}              & 2022 TPAMI    & 2D SOD                       & 69.56         & 229.10           & 0.045           & 0.631               & 0.652                 & \color{blue}{0.804}           & 0.798           & \color{blue}{0.790}            \\ \hline
				RCSB~\cite{Ke_2022_WACV}              & 2022 WACV     & 2D SOD                          & 27.25         & 454.23           & 0.067           & 0.590               & 0.488                & 0.675           & 0.755            & 0.652            \\ \hline
				ZoomNet~\cite{ZoomNet-CVPR2022}              & 2022 CVPR     & 2D SOD                       & 32.38         & 90.44           & \color{green}{0.039}            & \color{green}{0.712}               & \color{green}{0.689}                 & \color{green}{0.805}           & \color{green}{0.863}           & \color{green}{0.804}            \\ \hline
				TRACER~\cite{lee2021tracer}              & 2022 AAAI     & 2D SOD   & 3.90         & 2.78           & \color{blue}{0.044}           & \color{blue}{0.667}               & \color{blue}{0.648}                & 0.770           & \color{blue}{0.850}           & 0.740             \\ \hline \hline
				DDS~\cite{li2019distortion} & 2019 JSTSP    & $360^{\circ}$ SOD        & 27.23         & 60.36           & {0.045}           & {0.630}               & {0.635}                & {0.791}           & {0.808}           & {0.761}            \\ \hline 
				our	& 2022			& $360^{\circ}$ SOD                          & 56.86         & 42.87           &\color{red}{0.035}  & \color{red}{0.759}      & \color{red}{0.738}       & \color{red}{0.831}  & \color{red}{0.886}  & \color{red}{0.822}   \\ \hline
			\end{tabular}			
		}
	\end{table*}
	
	\subsubsection{Quantitative Evaluation}
	To demonstrate the effectiveness of the proposed method,  
	we selected the Top-5 in performance from 2D SOD methods with available training code in Tab.\ref{tab:benchmark}, i.e., PAKRN~\cite{xu2021locate}, PoolNet~\cite{Liu2019PoolSal}, RCSB~\cite{Ke_2022_WACV}, ZoomNet~\cite{ZoomNet-CVPR2022} and  TRACER~\cite{lee2021tracer}. 
	Then, for a fair comparison, we finetune these five models, DDS~\cite{li2019distortion} and our proposed model on the ODI-SOD train set. 
	After finetuning, their evaluation results on the ODI-SOD test set are listed in Tab.\ref{tab:PerformanceComp}. We can see that the overall metric scores are much better than those before finetuning in Tab.\ref{tab:benchmark} for the selected methods except for PAKRN~\cite{xu2021locate}. One primary reason may be that PAKRN~\cite{xu2021locate} needs multi-stage joint training, which is not easy for a new task. 
	Surpassing PAKRN~\cite{xu2021locate} and DDS~\cite{li2019distortion}, ZoomNet~\cite{ZoomNet-CVPR2022} becomes the best one, but the performance is nowhere near as good as on 2D SOD datasets, which illustrates the challenge of the proposed dataset again. Compared with the other methods, our method demonstrates sustained advantages and significant improvements on all the listed metrics and has become the new state-of-the-art. It verifies that the proposed method is effective for the $360^{\circ}$ ISOD task.
	
\begin{figure*}[!t] 
		\centering
		\subfloat[]{
			\begin{minipage}[t]{0.31\linewidth}
				\centering
				\includegraphics[width=0.95\columnwidth]{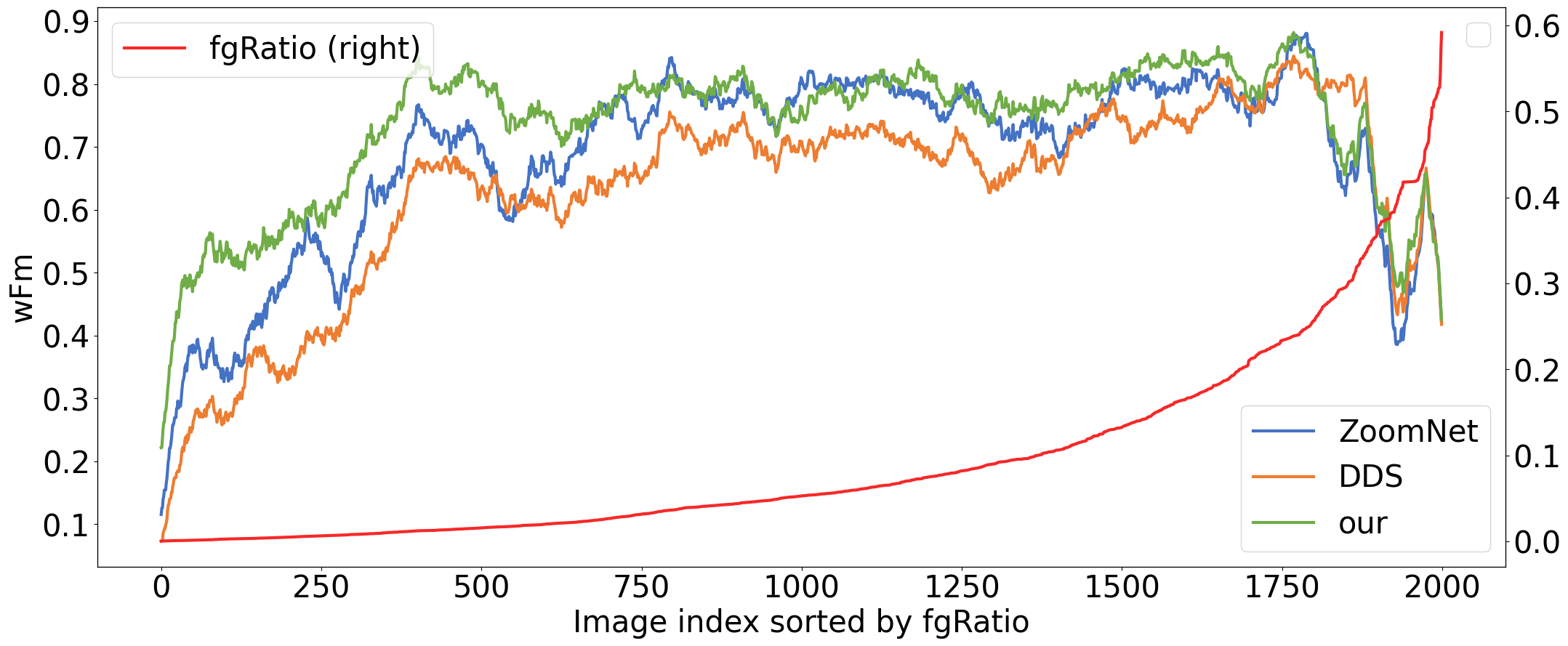}
			\end{minipage}
		}
		\subfloat[]{
			\begin{minipage}[t]{0.31\linewidth}
				\centering
				\includegraphics[width=0.95\columnwidth]{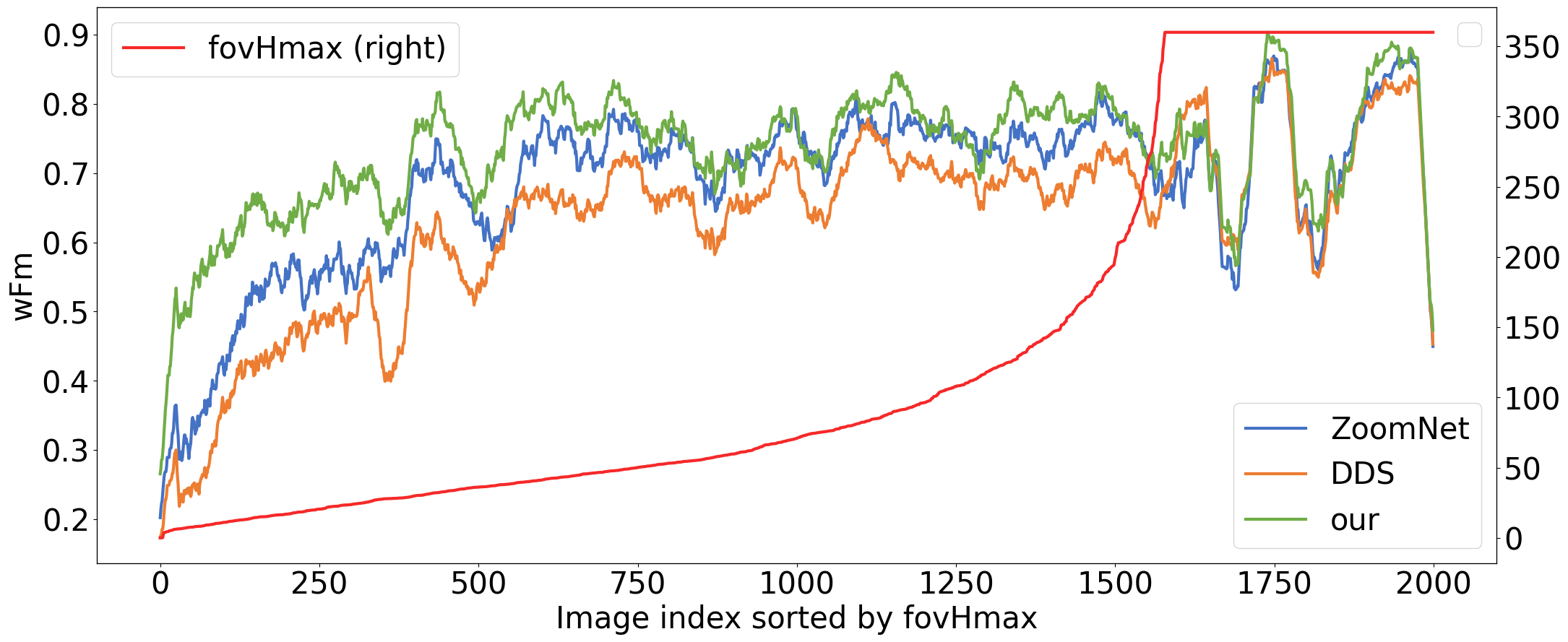}
			\end{minipage}
		}
		\subfloat[]{
			\begin{minipage}[t]{0.31\linewidth}
				\centering
				\includegraphics[width=0.95\columnwidth]{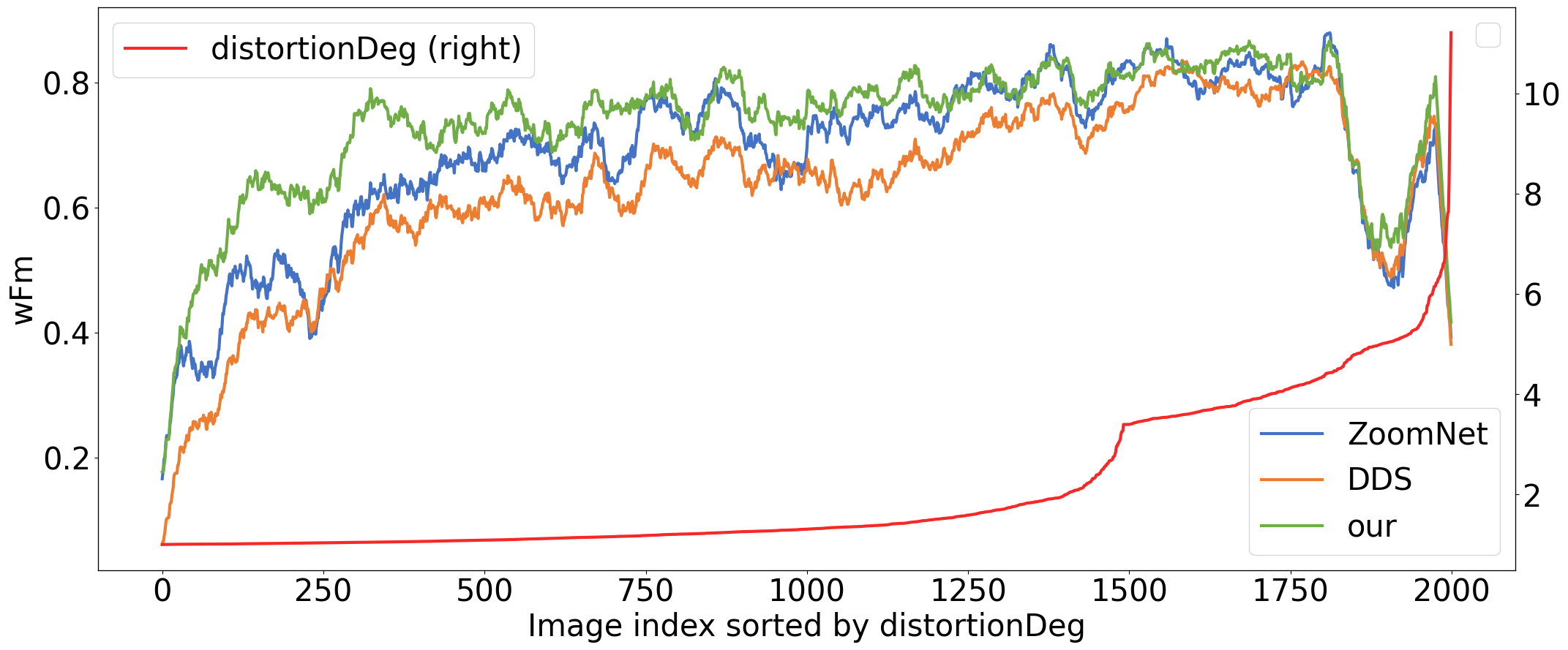}
			\end{minipage}
		}	
		\caption{Performance variation trends with sample attributes including (a) target foreground ratio, (b) max horizontal FoV coverage and (c) distortion degree. In each subfigure, the red line constitutes attribute values and the green line is our trends. }
		\label{fig:compare_by_attributes}
	\end{figure*}

	We sort the ODI-SOD test samples by different attributes, including the target foreground ratio, max horizontal FoV, distortion degree and discontinuous edge effect.
	To further evaluate the methods' performance variation trends with the attributes, 
	from the finetuned models in Tab.\ref{tab:PerformanceComp} we choose the best 2D model ZoomNet~\cite{ZoomNet-CVPR2022}, the 360$^{\circ}$ model DDS~\cite{li2019distortion} and our model to make predictions, statistics and analysis.
	Based on the predicted maps we compute the methods' $wF_{\beta}$ scores on each sample. Then, the statistic scores about discontinuous edge effects are shown in Tab.\ref{tab:w_wo_edge}, and the scores about the other attributes are plotted as broken line graphs in Fig.~\ref{fig:compare_by_attributes}.
	For better visualization and analysis, the lines are smoothed by a moving averaging window with size 50, and the secondary Y-axis is the attribute values.

	\begin{table}[]
		\caption{Performance on the ODI-SOD test subset with discontinuous edge effects. }
		\label{tab:w_wo_edge}
		\centering
		\begin{tabular}{l|cccccc}
			\hline
			\textbf{Methods} & MAE$\downarrow$ & $F_\beta$$\uparrow$ & $wF_\beta$$\uparrow$ & $S_m$$\uparrow$ & $E_m$$\uparrow$ & $maxF$$\uparrow$ \\ \hline			
			ZoomNet~\cite{ZoomNet-CVPR2022}          & 0.071           & 0.801               & 0.752                & 0.797           & 0.869    & 0.860   \\ \hline
			DDS~\cite{li2019distortion}              & 0.074           & 0.778               & 0.733                & 0.79           & 0.875  & 0.837      \\  \hline
			our              & 0.065           & 0.807               & 0.776                & 0.82           & 0.875   & 0.865        \\ \hline
		\end{tabular}		
	\end{table}

	From Tab.\ref{tab:w_wo_edge} we observe that our method outperforms other methods on all listed criteria and demonstrates significant advantages on MAE, $wF_\beta$ and $S_m$ measures. 
	It suggests that our method has fewer false predictions and better overall performance and that our prediction maps have more similar structures with ground-truth maps.  
	Overall, our method is effective for the discontinuous edge effects in panoramas.
	
	From Fig.\ref{fig:compare_by_attributes} we can notice that the performances of all the listed methods sharply decrease when processing intricate image samples such as those with very large/small target foreground area ratios, very wide/narrow FoV coverages and severe distortions. However, our method still presents consistent advantages for most image samples and the advantages become more and more evident as the target foreground area or FoV coverage gets smaller and smaller, which indicates that our method is better at processing small targets. On the whole, for samples with different attributes, our method performs better than other methods.

	\subsubsection{Qualitative Evaluation}
	\begin{figure*}[!t] 
		\centering
		\subfloat{
			\begin{minipage}[t]{0.85\linewidth}
				\centering
				\includegraphics[width=0.98\columnwidth]{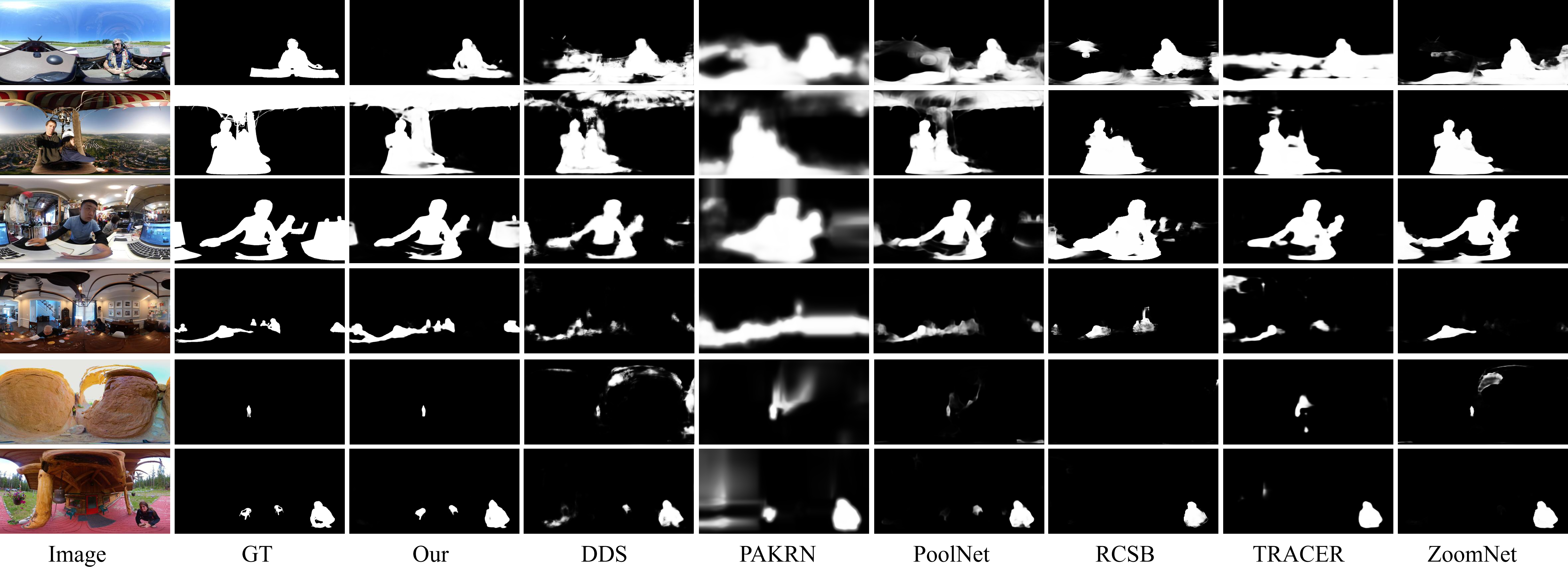}
			\end{minipage}
		}
		\caption{Visual testing examples of representative state-of-the-art algorithms after finetuning on the ODI-SOD train set. }
		\label{fig:benchmark_w_finetune}
	\end{figure*}

	Fig.\ref{fig:benchmark_w_finetune} shows some representative results of existing SOTA methods and our method on ODI-SOD test set. 
	We can perceive that most methods fail to segment severely distorted regions well, while our method has more robust adaptability to different distortions (e.g., the first two rows in Fig.\ref{fig:benchmark_w_finetune}).
	When processing the object with discontinuous edge effects especially one of the separated parts is very small, other methods usually ignore its part regions, in contrast, our method can segment it more completely (e.g., the middle two rows in Fig.\ref{fig:benchmark_w_finetune}). For changeable scale objects especially small objects, our method also outperforms other methods (e.g., the last two rows in Fig.\ref{fig:benchmark_w_finetune}).  Generally, our method can obtain better, more complete, more continuous and more uniform segmentation maps than other methods, which is effective for $360^{\circ}$ ISOD task.

	\subsubsection{Complexity Analysis}	
	In addition to the above evaluations, complexity analysis of the finetuned models is also conducted. We calculate the parameters and MACs (Multiply–Accumulate Operations) of the finetuned models and present the results in Tab.\ref{tab:PerformanceComp}. From Tab.\ref{tab:PerformanceComp}, we see that method TRACER~\cite{lee2021tracer} has the best parameters and MACs but relatively worse performance, and method ZoomNet~\cite{ZoomNet-CVPR2022} has fewer parameters but more MACs. In general, our method obtains a balance between model complexity and performance. For $512\times 256$ input images, our method runs at a speed of 7.6 FPS on one GTX 2080Ti GPU.
	


	\subsection{Ablation Study}
	
	To demonstrate the effectiveness of the proposed module SAVT with two sub-modules VT and SAF, we further conduct ablation studies on the ODI-SOD test set. The three branches in VT are also considered for detailed analysis. First, based on the baseline model, we add the gimped VT versions with only a single branch or two branches to test their effectiveness. Then, we try the complete submodel VT and SAF. The experimental results are shown in Tab.\ref{tab:Ablation}. 
	
	\begin{table*}[htp]
		\centering{
			\caption{Ablation study of our method on the ODI-SOD test set. Note: HRB, VRB and ZB are the horizontal rotation, vertical rotation and zooming branches in VT.}
			\label{tab:Ablation}
			\begin{tabular}{l|cccc|cccccc}
				\hline
				\textbf{Method}  & \textbf{HRB} & \textbf{VRB} & \textbf{ZB} & \textbf{SAF} & MAE$\downarrow$ & $F_\beta$$\uparrow$ & $wF_\beta$$\uparrow$ & $S_m$$\uparrow$ & $E_m$$\uparrow$ & $maxF$$\uparrow$ \\ \hline
				Baseline         &              &              &             &              & 0.04            & 0.73                & 0.713                & 0.817           & 0.875           & 0.803              \\ 
				Baseline+VT\_H     & \checkmark   &              &             &              & 0.037           & 0.746               & 0.728                & 0.826           & 0.88            & 0.807            \\ 
				Baseline+VT\_V     &              & \checkmark   &             &              & 0.038           & 0.739               & 0.719                & 0.822           & 0.876           & 0.805            \\ 
				Baseline+VT\_Z      &              &              & \checkmark  &              & 0.038           & 0.742               & 0.725                & 0.824           & 0.878           & 0.811            \\ 
				Baseline+VT\_HV & \checkmark   & \checkmark   &             &              & 0.038           & 0.751               & 0.728                & 0.822           & 0.879           & 0.818            \\ 
				Baseline+VT\_HZ  & \checkmark   &              & \checkmark  &              & 0.037           & 0.751               & 0.731                & 0.827           & 0.88            & 0.816            \\ 
				Baseline+VT\_VZ  &              & \checkmark   & \checkmark  &              & 0.037           & 0.741               & 0.727                & 0.824           & 0.877           & 0.813            \\ 
				Baseline+VT      & \checkmark   & \checkmark   & \checkmark  &              & 0.038           & 0.75                & 0.734                & 0.83            & 0.882           & 0.818            \\ \hline
				Baseline+SAVT(VT+SAF)  & \checkmark   & \checkmark   & \checkmark  & \checkmark   & \color{red}{0.035}  & \color{red}{0.759}      & \color{red}{0.738}       & \color{red}{0.831}  & \color{red}{0.886}  & \color{red}{0.822}   \\ \hline
			\end{tabular}
		}
	\end{table*}

	\subsubsection{Effectiveness of VT}
	From the first four rows in Tab.\ref{tab:Ablation} we observe that all the performances can be improved when only using one transform branch, especially the versions VT\_H and VT\_Z, which indicates the single transform branch is effective. When randomly adopting two transform branches in VT, most metrics get better than those using only one, suggesting that the models with two transform branches still work well. When utilizing the complete VT, the overall performance is further enhanced. Compared with the baseline, the MAE score becomes 0.038 from 0.040 and the $F_\beta$ increases to 0.759 from 0.730. It verifies the sub-model VT is effective. It is worth mentioning that more transform branches mean more diversities of features. This requires a stronger feature fusion operation to obtain the desired features. Next, we will further verify the ability of SAVT to fuse different types of transformed features.

	\begin{table}[!h]
		\caption{Influence of rotation degrees in the horizontal/vertical rotation branch (HRB/VRB) and zooming scale factors in the zooming branch (ZB). For HRB, degrees ranges from $-180^{\circ}$ to $180^{\circ}$ except $0^{\circ}$.}
		\begin{tabular}{c|c|c|c|c|c|c}
			\hline
			\textbf{Parameters} & MAE$\downarrow$ & $F_\beta$$\uparrow$ & $wF_\beta$$\uparrow$ & $S_m$$\uparrow$ & $E_m$$\uparrow$ & $maxF$$\uparrow$ \\ \hline
			Degree step			& \multicolumn{6}{c}{HRB}  \\ \hline
			30$^{\circ}$        & 0.035           & 0.758               & 0.738                & 0.829           & 0.885           & 0.823            \\ \hline
			45$^{\circ}$        & 0.037           & 0.754               & 0.743                & 0.833           & 0.884           & 0.816            \\ \hline
			60$^{\circ}$        & 0.039           & 0.748               & 0.738                & 0.828           & 0.879           & 0.816            \\ \hline
			\hline
			Degrees  			& \multicolumn{6}{c}{VRB}  \\ \hline
			[$\pm30^{\circ}$]                               & 0.035           & 0.758               & 0.738                & 0.829           & 0.885           & 0.823            \\ \hline
			{[}$\pm60^{\circ},\pm30^{\circ}${]} & 0.039           & 0.745               & 0.729                & 0.827           & 0.876           & 0.814            \\ \hline
			{[}$\pm45^{\circ},\pm30^{\circ}${]} & 0.039           & 0.745               & 0.727                & 0.826           & 0.878           & 0.812            \\ \hline
			\hline
			Scale factors  		& \multicolumn{6}{c}{ZB}  \\ \hline
			{[}0.8,1.2,0.7,1.3{]} & 0.035           & 0.758               & 0.738                & 0.829           & 0.885           & 0.823            \\ \hline
			{[}0.8,1.2,0.6,1.4{]} & 0.038           & 0.749               & 0.731                & 0.826           & 0.88            & 0.814            \\ \hline
			{[}0.7,1.3,0.5,1.5{]} & 0.039           & 0.747               & 0.726                & 0.823           & 0.88            & 0.817            \\ \hline
		\end{tabular}
		
		\label{tab:branch_parameter}
	\end{table}

	\subsubsection{Effectiveness of SAF}
	SAF is based on VT to assist subsequent feature fusion by adaptively adjusting the weights of different types of transformed features. From Tab.\ref{tab:Ablation} we find that all the metrics are improved after using SAF. After adding SAF, The MAE score becomes 0.035 from 0.038 and the $F_{\beta}$ score becomes 0.759 from 0.750, which presents the effectiveness of SAF.
	
	Overall, the performance is significantly improved from the baseline to our final model with the proposed SAVT. As shown in Tab.\ref{tab:Ablation}, the MAE score decreases to 0.035 from 0.040 and the $F_{\beta}$ score increases to 0.759 from 0.730. It shows the proposed module SAVT is very effective for the $360^{\circ}$ ISOD task.

	\subsubsection{Influence of Parameters in VT}
	To investigate the influence of transform branches' parameters in VT, 
	We try several groups of parameters about the rotation degrees of the horizontal/vertical branch and the scale factors of the zooming branch based on the final model. 
	From the results in Tab.\ref{tab:branch_parameter} we find that for HRB the overall performance decreases a little when the rotation degree becomes $45^{\circ}$ from $30^{\circ}$, and when the degree becomes $60^{\circ}$ the performance is more negatively affected. No smaller degree step is tried as the angular resolution is enough for our task (e.g., 1.42 pixels per degree for a 512$\times$256 ERP image). 
	For VRB, we choose $\pm30^{\circ}$ since the extra larger degrees cannot bring better performance.
	It is reasonable and realistic as the vertical field of view range is $\pm90^{\circ}$ and we usually do not look up or down too much. Moreover, the ERP feature appearance is more sensitive to vertical rotation than horizontal rotation. The inappropriate larger degrees may bring overly deformed appearance and unexpected distractors, which is not beneficial for feature perception and performance gains.
	As for ZB, the first group parameters are best. Both smaller and bigger scale factors are not suitable. It is also realistic as excessive zooming in/out cannot help to learn better features and appropriate transformations are more important. 


	\subsection{Failure Cases}	
	\begin{figure}[!t] 
		\centering
		\subfloat{
			\begin{minipage}[t]{0.98\linewidth}
				\centering
				\includegraphics[width=0.98\columnwidth]{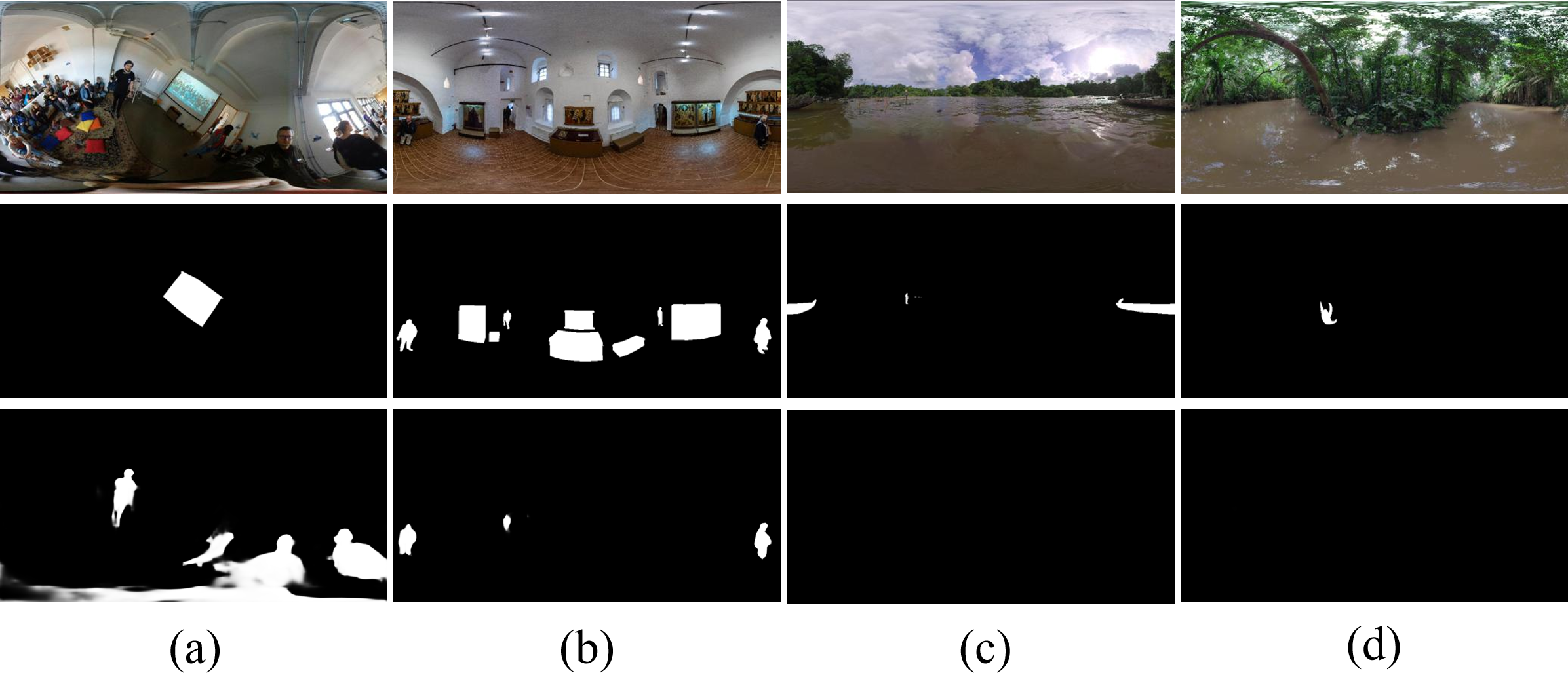}
			\end{minipage}
		}
		\caption{Failure cases of our method. Rows from top to bottom: images, ground-truth masks and our results. }
		\label{fig:failureCases}
	\end{figure}
	Beyond the successful cases, we show representative failure cases in Fig.\ref{fig:failureCases}.
	 We find when the scene contains many objects or the target object is camouflaged, our approach is more prone to failure. For example, the scenes in Fig.\ref{fig:failureCases}(a) and Fig.\ref{fig:failureCases}(b) contains many objects while our method cannot detect the targets or miss some targets. To some extent, it is caused by the way of defining salient objects as it is hard to define salient targets in panoramic scenes with many objects, especially when the objects have similar saliency. Besides, when target objects are similar with background or camouflaged in background (e.g., the boat in Fig.\ref{fig:failureCases}(c) or the animal in Fig.\ref{fig:failureCases}(d)), our method also fails to find the targets.

	\subsection{Discussion}
	Although our method is effective and outperforms the SOTA methods, it has three limitations. 
	Firstly, the adopted resolution in the study is not large for panoramas with wide FoV and rich information. Small resolutions can lose some important details. 
	Secondly, to use the mature 2D CNNs, the original spherical images must be projected onto the 2D plane, resulting in different degrees of geometrical distortion. Thirdly, the model is not lightweight and efficient enough. 
	Therefore, how to make full use of the original image information and investigate a highly efficient solution need to be further explored in the future. 
	In addition to methods, some special scenes can also be explored. For example, the target objects in the scene sometimes are concealed or camouflaged, “seamlessly” embedded in their surroundings~\cite{fan2021concealed, fan2020camouflaged}, or the objects are in the clutter~\cite{fan2018salient, fan2022salient}, which are challenging situations for panoramic scenarios and should be further discussed in future work.

	\section{Conclusion}
	In this paper, we construct a $360^{\circ}$ omnidirectional image-based SOD dataset, namely ODI-SOD, to explore the salient object detection in panoramic scenes. It has object-level pixel-wise annotations on ERP images and is the largest dataset for $360^{\circ}$ ISOD by far to our best knowledge. 	
	Moreover, inspired by humans' observing process, we propose a view-aware salient object detection method for $360^\circ$ ODIs, containing a novel module SAVT with two submodels VT and SAF. 
	VT is designed to simulate the process of looking left and right, up and down, far and near by changing the viewpoint or view distance. 
	SAF aims to adaptively fuse the output features of transform branches and the original learning branch based on different input samples. It can flexibly adjust the weights of different transformed features and obtain better fusion features. 
	Integrated SAVT effectively mitigates the effects of diverse distortion degrees, discontinuous edge effects and changeable object scales.
	Furthermore, we conduct qualitative and quantitative experiments to explore the proposed method and verify its effectiveness.

	\section*{Acknowledgments}
	This work is partially supported by the National Natural Science Foundation of China under Grant 62132002 and Grant 62102206.

	\bibliographystyle{IEEEtran}
	\bibliography{360SOD}

	\newpage
	
	\vspace{11pt}
	

	\vfill
	
\end{document}